\documentclass[runningheads]{llncs}

\usepackage{eccv}

\usepackage{eccvabbrv}

\usepackage{graphicx}
\usepackage{animate}
\usepackage{booktabs}

\usepackage[accsupp]{axessibility}  
\usepackage{tabularx, makecell, booktabs, adjustbox}
\usepackage{xcolor}
\usepackage{color, colortbl}
\usepackage[font=small, skip=4pt]{caption}

\usepackage{hyperref}

\usepackage{orcidlink}

\newcommand{\xb}{{\boldsymbol x}}

\newcommand{\yb}{{\boldsymbol y}}

\newcommand{\cb}{{\boldsymbol c}}

\newcommand{\x}{{\boldsymbol x}}

\usepackage{xparse}
\usepackage{xcolor}

\NewDocumentCommand{\todo}{o +m}{%
  \IfValueTF{#1}{%
    \IfEqCase{#1}{%
      {original}{\textcolor{blue}{#2}}%
      {changed}{\textcolor{red}{#2}}%
    }[\color{red}{#2}]%
  }{%
    \color{red}{#2}%
  }%
}
\newcommand{\name}{Memory-V2V\xspace}

\usepackage{xspace}
\usepackage{algorithm2e}
\usepackage{algorithmic}

\begin{document}

\title{Memory-V2V: Memory-Augmented Video-to-Video Diffusion for Consistent Multi-Turn Editing}

\titlerunning{Memory-V2V}

\author{Dohun Lee\inst{1,2,}\thanks{Work done during an internship at Adobe Research.}\orcidlink{0009-0005-2622-5595} \and
Chun-Hao Paul Huang\inst{1}\orcidlink{0000-0002-1268-6527} \and
Xuelin Chen\inst{1}\orcidlink{0009-0007-0158-9469} \and
Jong Chul Ye\inst{2}\orcidlink{0000-0001-9763-9609} \and
Duygu Ceylan\inst{1}\orcidlink{0000-0002-2307-9052} \and
Hyeonho Jeong\inst{1}\orcidlink{0000-0002-6864-4190}}

\authorrunning{D.~Lee et al.}

\institute{Adobe Research, London, UK \\ 
\email{\{chunhaoh, xuelinc, ceylan, hyeonhoj\}@adobe.com} \and
Kim Jaechul Graduate School of AI, KAIST, Daejeon, South Korea \\
\email{\{leedh7, jong.ye\}@kaist.ac.kr}\\
\url{https://dohunlee1.github.io/MemoryV2V/}}

\maketitle

\begin{center}
    \centering
    \captionsetup{type=figure}
    \vspace{-3mm}
    \includegraphics[width=\textwidth]{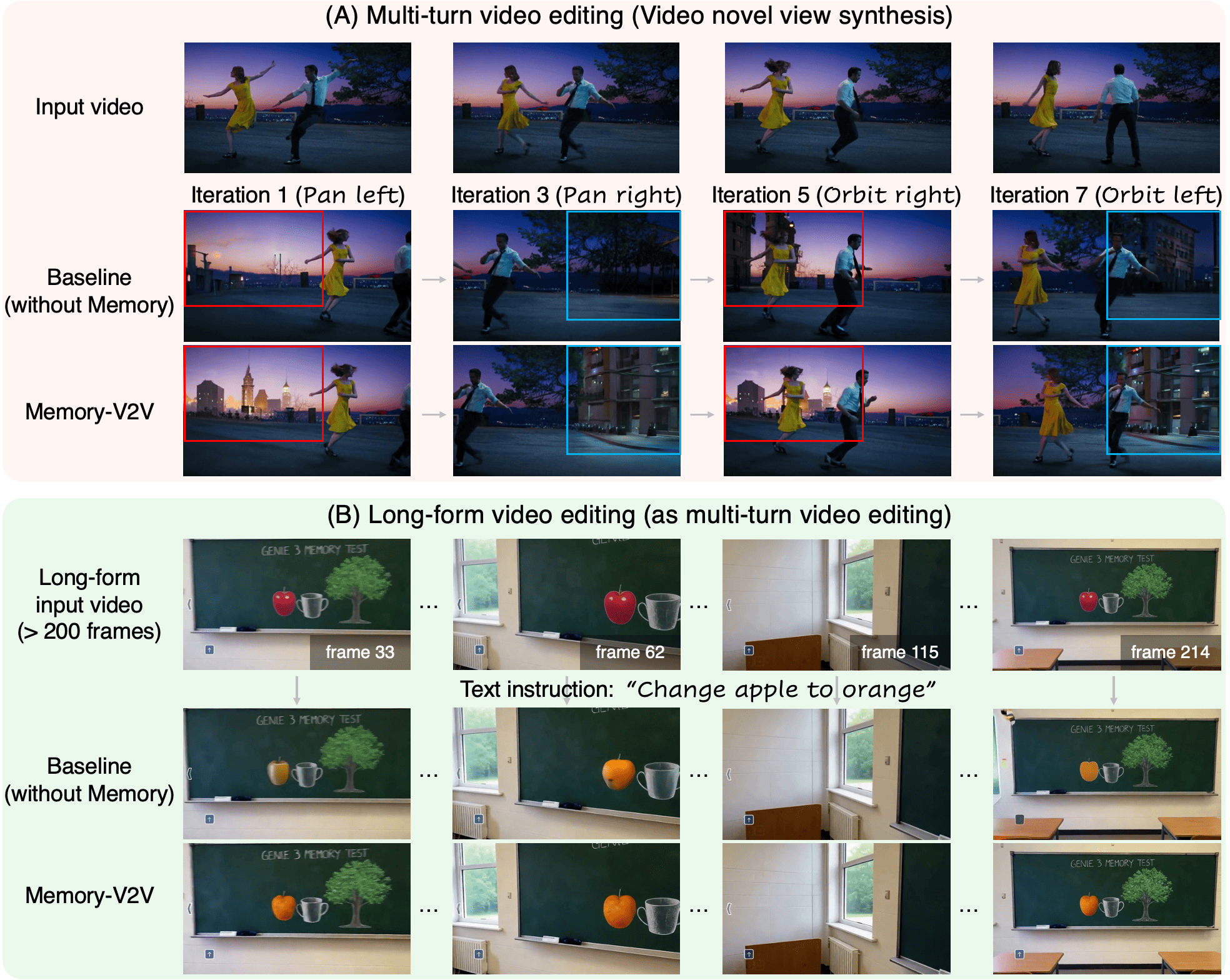}
    \captionof{figure}{
    \textbf{\name enables iterative video editing with long-term memory, producing results that remain consistent with edits from previous turns.}
    \textbf{(A)} and \textbf{(B)} show results for video novel view synthesis and text-guided long video editing, respectively.
    Colored boxes in \textbf{(A)} highlight novel-view regions that are expected to remain consistent across editing iterations.
    Each iteration performs independent denoising.
    Additional video results are provided on the supplementary project page.
    }
    \vspace{-2mm}
    \label{figure: teaser}
\end{center}%

\begin{abstract}
Video-to-video diffusion models achieve impressive single-turn editing performance, but practical editing workflows are inherently iterative. When edits are applied sequentially, existing models treat each turn independently, often causing previously generated regions to drift or be overwritten. We identify this failure mode as the problem of cross-turn consistency in multi-turn video editing.
We introduce \name, a memory-augmented framework that treats prior edits as structured constraints for subsequent generations. \name maintains an external memory of previous outputs, retrieves task-relevant edits, and integrates them through relevance-aware tokenization and adaptive compression. These technical ingredients enable scalable conditioning without linear growth in computation.
We demonstrate \name on iterative video novel view synthesis and text-guided long video editing. \name substantially enhances cross-turn consistency while maintaining visual quality, outperforming strong baselines with modest overhead.

\end{abstract}

\section{Introduction}
\label{sec: introduction}

Videos are rapidly becoming a dominant medium of modern communication and expression, spanning domains from entertainment to robotics simulation. With the advent of large-scale generative models, users increasingly expect video editing tools that enable attribute-level control — allowing modification of subjects~\cite{jiang2025vace, LucyEdit, ju2025editverse}, motion~\cite{jiang2025vace, LucyEdit, jeong2024vmc}, or viewpoint~\cite{bai2025recammaster, yu2025trajectorycrafter, ren2025gen3c}. However, in practice, video editing workflows are inherently iterative: users progressively refine outputs over multiple interactions, a setting we term \emph{multi-turn video editing}. 

While recent video editing frameworks demonstrate impressive results under single-pass interaction, they often fail to maintain cross-turn consistency across sequential edits.
For example, as shown in Fig.\ref{figure: teaser}(A), a state-of-the-art video novel view synthesis model, ReCamMaster \cite{bai2025recammaster}, successfully re-renders the input video from multiple target viewpoints, yet the novel-view regions across iterations remain inconsistent.
Similarly, in Fig.\ref{figure: teaser}(B), when segments of a long video are iteratively edited using LucyEdit \cite{LucyEdit}, each segment adheres to the local editing prompt, but the global appearance 
gradually drifts. For instance, an apple edited into an orange appears as visually different oranges across segments.

In this work, we formulate consistent multi-turn video editing as a distinct problem setting and introduce \name, a memory-augmented framework tailored to this scenario. Our central design principle is that multi-turn editing requires the memory to function as a constraint mechanism. This is fundamentally different from existing memory-augmented long video generators that assume temporal continuity. Hence, instead of treating previous outputs as additional temporal context, we regard them as a structured external memory that selectively enforces consistency across editing iterations.

A naive solution for enforcing multi-turn consistency is to condition on all previous edits. However, this leads to prohibitive computational growth and introduces redundant or irrelevant context. Instead, we introduce \name guided by two design principles: selective retrieval and relevance-aware capacity allocation. First, rather than conditioning on all previous edits or temporal adjacency retrieval used in autoregressive generation, we retrieve only the most relevant prior videos for the current turn. Crucially, retrieval is task-aware and non-temporal. For video novel view synthesis, we introduce a geometric Field-of-View (FOV) retrieval mechanism that ranks prior edits based on camera overlap. For text-guided long video editing, we retrieve segments according to semantic similarity of their corresponding source videos. Second, we allocate computational capacity proportionally to memory relevance. Retrieved videos are dynamically tokenized with varying spatial resolutions, preserving fine-grained detail for highly relevant edits while aggressively compressing less relevant ones. Furthermore, we introduce adaptive token merging based on attention responsiveness, reducing redundant computation while maintaining essential visual cues. Together, these components enable scalable conditioning on heterogeneous memory sources without linear growth in cost.

We evaluate \name on two representative video-to-video editing tasks: video novel view synthesis and text-guided long video editing. 
\name substantially improves cross-iteration consistency in both tasks while also outperforming state-of-the-art baselines in single-turn editing performance.
Our contributions can be summarized as follows:

\begin{itemize}
    \item We formulate \textit{multi-turn video editing} as a distinct problem setting, highlighting the challenge of preserving consistency across independently denoised edits.
    \item We propose \name, a memory augmented framework for iterative editing, where prior edits are treated as structured constraints rather than temporal continuations. 
    \item We introduce task-aware retrieval mechanisms, including FOV-based geometric retrieval and semantic segment retrieval, that select relevant prior edits without incurring linear context growth. We develop dynamic tokenization and adaptive token merging enabling scalable multi-turn conditioning with reduced computational overhead.
    \item 
    We demonstrate substantial improvements in cross-turn consistency on iterative novel view synthesis and long video editing. To enable training for long video editing, we introduce a dataset augmentation pipeline that temporally extends only target videos from short source–target pairs, eliminating the need for long paired video-to-video training data.
\end{itemize}

\section{Related Work}
\label{sec: related work}

\subsubsection{Video novel view synthesis.}
Given a monocular video of a dynamic scene, video novel view synthesis aims to generate plausible videos captured from unseen camera trajectories, while preserving the underlying 3D structure and temporal dynamics \cite{van2024generative}.
Due to the scarcity of large-scale paired multi-view video datasets, prior works have focused on test-time optimization or scene-specific overfitting of existing monocular video generators \cite{mengnvs, zhang2025recapture, jeong2025reangle}.
Recently, approaches such as ReCamMaster \cite{bai2025recammaster} leverage large-scale synthetic 4D datasets rendered from simulation engines \cite{greff2022kubric, engine2018unreal} to finetune text-to-video diffusion models into multi-view video-to-video models. As shown in Fig.~\ref{figure: teaser}(A), when such models are applied iteratively, the generated outputs fail to maintain consistency for regions that were not visible in the input video. In contrast, as demonstrated in Fig.~\ref{figure: teaser}(A), we incorporate an explicit visual memory into pre-trained models such as ReCamMaster~\cite{bai2025recammaster} to ensure strong cross-iteration consistency.

Another line of work \cite{ren2025gen3c, yu2025trajectorycrafter, wang2025epic, jeong2025reangle} employs point-cloud renderings, which, though incomplete, approximate the desired camera trajectories as geometric proxies. Video diffusion models then use these renderings as spatial guidance to synthesize complete and geometrically consistent videos.
While some works iteratively refine point clouds to improve static scene novel view synthesis \cite{ren2025gen3c, yu2024viewcrafter}, they fail to generalize to dynamic scenes with non-rigid motion.

\subsubsection{Text-guided video editing.}
Early text-driven video editing approaches can be broadly divided into two categories \cite{LucyEdit}.
The first relies on inference-time solutions, often using deterministic inversion with test-time optimization to align edits with text prompts \cite{wu2023tune, ceylan2023pix2video, geyer2023tokenflow, jeong2024vmc}.
The second conditions video diffusion models on explicit visual cues such as depth maps or optical flow fields \cite{chen2023control, jeong2023ground, cong2023flatten, gu2024videoswap}.
While effective in constrained settings, they often introduce temporal artifacts and struggle to generalize to flexible, open-domain editing scenarios.

Recently, instruction-based foundational video-to-video editing models \cite{LucyEdit, jiang2025vace, qin2024instructvid2vid, polyak2024movie, ju2025editverse} have emerged, jointly conditioning on source videos and text instructions.
These models deliver precise, high-fidelity edits while preserving subject identity and motion.
However, these models are constrained by a limited temporal context window. A naive workaround is to split a long video into shorter segments and edit each independently, but as shown in Fig.~\ref{figure: teaser}(b), this approach leads to significant appearance inconsistencies across segments. 
In contrast, as illustrated in Fig.~\ref{figure: teaser}(b), our memory mechanism enables coherent long-form video editing with consistent appearance and motion across segments. By casting long video editing as a multi-turn editing problem and equipping video-to-video diffusion models~\cite{LucyEdit} with explicit visual memory, we achieve precise and temporally consistent results, \textit{even when segments are denoised independently.}

\subsubsection{Long video generation with memory or context.}
Although modern video models generate visually impressive results, they inherently lack 3D and long-term consistency due to their 2D frame-based representations and limited temporal context \cite{team2025hunyuanworld}. 
To address this, recent works first reformulate full-sequence video diffusion models into auto-regressive generators \cite{ruhe2402rolling, chen2024diffusion, huang2025self, yin2024slow}, then incorporate memory modules that capture information from past generations.
One line of work maintains an external cache of previous generations and conditions the current generation by retrieving a subset of them, either as raw RGB frames \cite{li2025vmem, xiao2025worldmem, yu2025context} or as hidden states \cite{cai2025mixture, yu2025malt}.
Alternatively, other methods aim to compress the long-term context into compact forms such as latent states \cite{dalal2025one, po2025long, zhang2025test}, semantic video tokens \cite{ouyang2025tokensgen, jiang2025lovic, zhang2025packing}, or vision–language model features \cite{chen2025longanimation}. Unlike prior literature, we address memory across independently generated editing iterations. In this setup, the model must selectively maintain consistency across multiple edited videos rather than a continuous frame stream.
\begin{figure*}[!t]
    \centering
    \includegraphics[width=\textwidth]{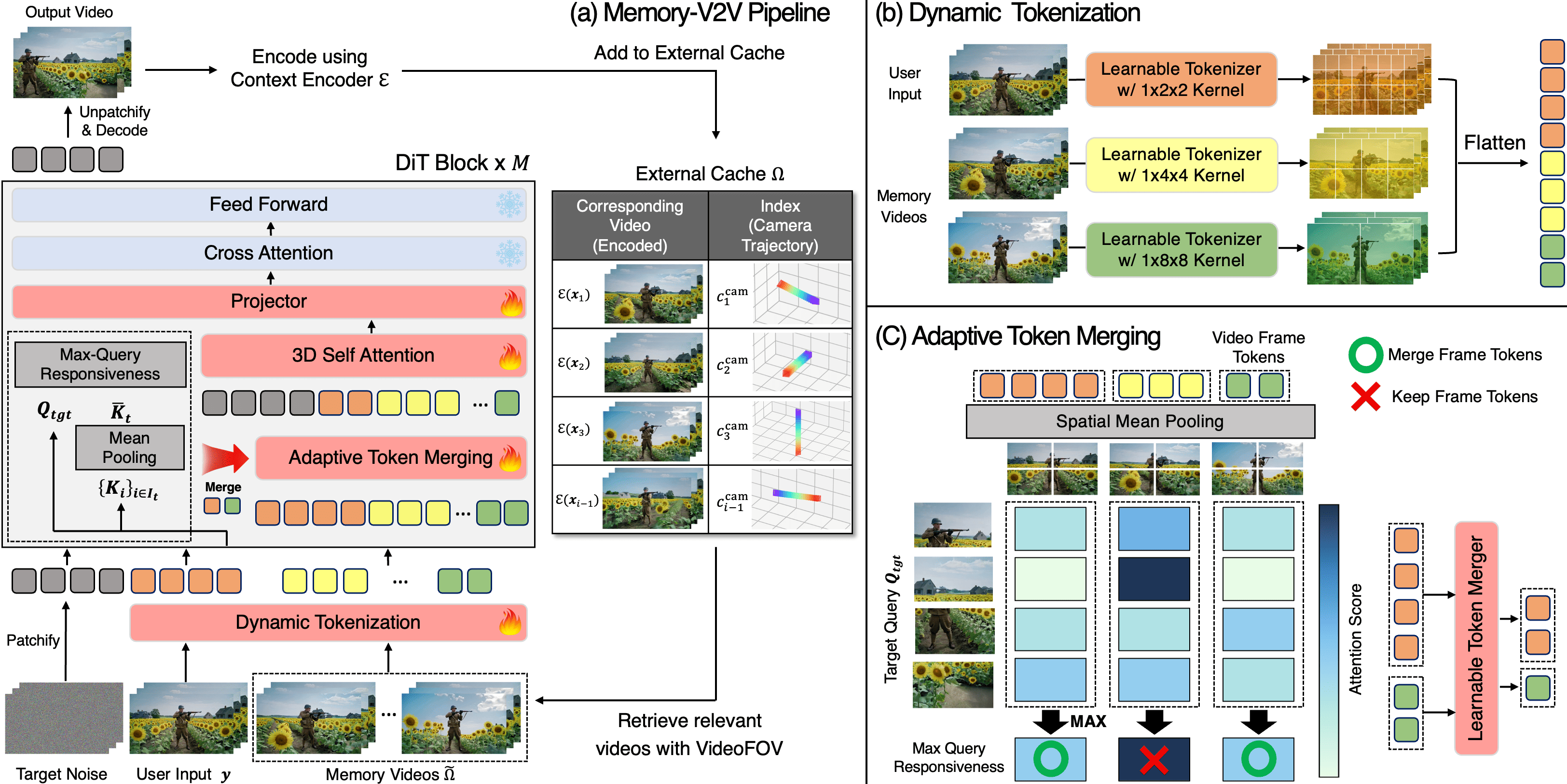}
    \caption{
    \textbf{Memory-V2V framework.}
    (a) From an external cache of previously edited videos, only the top-$k$ most relevant videos are retrieved and used as memory inputs to ensure cross-iteration consistency.
    (b) Dynamic tokenizers allocate more tokens to highly relevant videos—preserving fine details while maintaining an efficient overall token budget.
    (c) Adaptive token merging further reduces redundant computations by compressing less informative frames based on their attention-based responsiveness to the target query.}
    \label{fig: overview}
\end{figure*}

\section{Memory-V2V}
\label{sec: method}

\subsection{Overview}

Existing video-to-video diffusion models are \textit{single-turn video editing} models, which are trained to generate an output video conditioned on a single source video and auxiliary editing signals, approximating the conditional distribution $p(\xb \, | \, \yb, \, \cb)$,
where $\xb$ is the output video, $\yb$ is the user-provided source video, and $\cb$ denotes additional conditions like text or camera pose.
In contrast, our \textit{multi-turn video editing} task requires incorporating prior editing history by allowing $\yb$ to include both the current input video and previously edited videos, enabling the model to maintain cross-edit consistency across iterations.

Memory-V2V achieves this through a hybrid retrieval and compression strategy, built upon an effective memory representation. In the following sections, we first detail our framework in the context of video novel view synthesis.
We consider ReCamMaster \cite{bai2025recammaster}, a pretrained video DiT model capable of single-turn video novel view synthesis, as our base model. We then introduce the core components of \name, including an efficient video retrieval mechanism that selectively fetches relevant past videos from an external cache, dynamic tokenization strategies to represent the retrieved videos, and a compressor that effectively processes the retrieved information (Sec.~\ref{sec: method-token-merging}). Retrieval is non-parametric, while the tokenization and merging modules are learned end-to-end. Finally, we extend Memory-V2V to text-guided, long-form video editing by formulating it as a multi-turn video editing task (Sec.~\ref{sec: method-long-video-editing}).

\subsection{Video Retrieval-based Dynamic Tokenization}
\label{sec: method-retrieval-tokenization}
After each editing iteration $j$, we store the latent video $\mathcal{E}(\xb_{j}) \in \mathbb{R}^{F \times H \times W \times C}$ in an external cache $\Omega$, indexed by its corresponding camera trajectory $\cb_{j}^\text{cam}$. 
The latent video's temporal and spatial dimensions are denoted by $F,H,W$, and $C$ is the latent channel size, and the context encoder $\mathcal{E}$ corresponds to the VAE encoder.
At the $i$-th editing iteration ($i {>} j {>} 0$), our goal is to produce a new video $\xb_{i}$ that remains consistent with both the user-input video $\yb$ and the previously generated videos in $\Omega = \{ (\mathcal{E}(\xb_{j}), \cb_{j}^\text{cam}) \, {\mid} \, 0 < j < i \}$, while accurately following the new target camera trajectory $\cb_{i}^\text{cam}$. Since the total number of cached frames in $\Omega$ increases rapidly over multiple editing turns, it is infeasible to condition the pretrained video-to-video DiT model on the entire editing history.
We argue that only a subset of the previously edited videos is necessary to maintain spatial and temporal consistency.
To this end, we sort all cached latent videos in $\Omega$ and dynamically retrieve a set of the top-$k$ most relevant ones, denoted as $\tilde{\Omega}$.

\subsubsection{Video FOV retrieval.}
In case of video novel-view synthesis, we determine the relevance via our proposed VideoFOV retrieval algorithm, which quantifies the geometric overlap between the field-of-view (FOV) of the target camera trajectory $\cb_{i}^\text{cam}$ and those of cached videos.
Specifically, given a target camera trajectory $\cb_{i}^\text{cam} = \{ \cb_{i,t}^\text{cam} \, {\mid} \, 1 \leq t \leq F \}$, where $F$ is the number of frames, we place a unit sphere at the first camera position $\cb_{i,1}^\text{cam}$ and uniformly sample $M$ viewing directions on its surface \footnote{In all experiments, we set $M=64,800$.}.
For each frame along the trajectory, a sampled point is marked as visible if it lies within the projected image bounds and has positive depth after being transformed to the camera’s local coordinate system.
The per-frame FOV $\mathcal{F}_\text{frame}(\cb_{i,t}^\text{cam})$ is defined as the set of visible sampled points, and the video-level FOV $\mathcal{F}_\text{video}(\cb_{i}^\text{cam})$ is obtained as the union of all frame-level FOVs:
\begin{equation}
\mathcal{F}_\text{video}(\cb_{i}^\text{cam}) = \bigcup_{t=1}^{F} \mathcal{F}_\text{frame}(\cb_{i,t}^\text{cam})  \, .
\label{eq:fov_1}
\end{equation}

Given a target trajectory $\cb_{i}^\text{cam}$ and a cached trajectory $\cb_{j}^\text{cam}$, we define two complementary FOV similarity metrics:
\begin{equation}
\begin{split}
s_{\text{overlap}}(\cb_{i}^\text{cam}, \cb_{j}^\text{cam})
=
\frac{\lvert \mathcal{F}_\text{video}(\cb_{i}^\text{cam}) \cap \mathcal{F}_\text{video}(\cb_{j}^\text{cam}) \rvert}
     {\lvert \mathcal{F}_\text{video}(\cb_{i}^\text{cam}) \cup \mathcal{F}_\text{video}(\cb_{j}^\text{cam}) \rvert},
\\
s_{\text{contain}}(\cb_{i}^\text{cam}, \cb_{j}^\text{cam})
=
\frac{\lvert \mathcal{F}_\text{video}(\cb_{i}^\text{cam}) \cap \mathcal{F}_\text{video}(\cb_{j}^\text{cam}) \rvert}
     {\lvert \mathcal{F}_\text{video}(\cb_{i}^\text{cam}) \rvert}.
\end{split}
\label{eq:fov_2}
\end{equation}

The final relevance score is a weighted combination:
\begin{equation}
s(\cb_{i}^\text{cam}, \cb_{j}^\text{cam})
=
\lambda \, s_{\text{overlap}}(\cb_{i}^\text{cam}, \cb_{j}^\text{cam})
+
(1 - \lambda) \, s_{\text{contain}}(\cb_{i}^\text{cam}, \cb_{j}^\text{cam}),
\label{eq:fov_3}
\end{equation}
where we set $\lambda{=}0.5$.

All cached videos in $\Omega$ are then ranked by $s(\cb_{i}^\text{cam}, \cb_{j}^\text{cam})$, and the top-$k$ highest-scoring videos are selected as the retrieved video set $\tilde{\Omega}$.
The full algorithm and implementation details are provided in the supplementary material.

\begin{figure*}[!t]
    \centering
    \includegraphics[width=\textwidth]{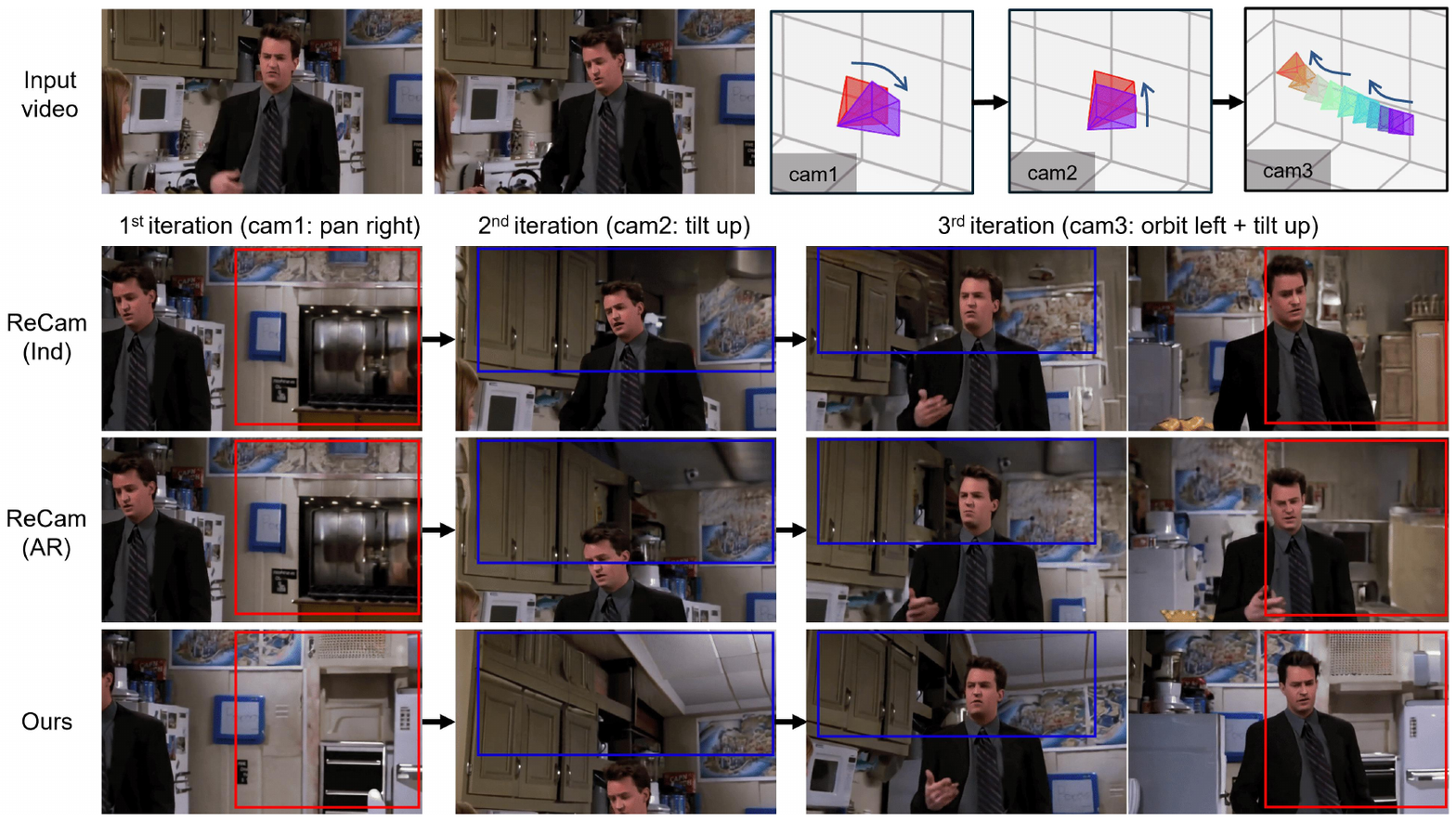}
    \vspace{-12mm}
    \caption{
    \textbf{Qualitative results for multi-turn video novel view synthesis}.
    Compared to baselines, Memory-V2V (Ours) generates videos from new camera trajectories while maintaining consistency across all novel regions generated (e.g., red \& blue box).
    }
    \label{fig: novel-view-synthesis-comparison}
\end{figure*}

\subsubsection{Dynamic tokenization.}
Unlike long video generation \cite{xiao2025worldmem, li2025vmem}, where the memory context typically consists of a small number of frames (${<} 10$), the multi-turn video editing treats entire videos as retrieval units.
As iterations accumulate, the total number of past conditioning frames can easily reach hundreds \footnote{For video novel view synthesis, a single video contains 81 frames.}, making direct encoding computationally prohibitive.
Hence, we propose to tokenize the conditional videos dynamically based on their relevance to the current edit.

Our goal is to apply multiple learnable tokenizers to each retrieved video.
In practice, we train tokenizers with spatio-temporal compression factors ($f'{\times}h'{\times}w'$) of $1{\times}2{\times}2$, $1{\times}4{\times}4$, and $1{\times}8{\times}8$, where the $1{\times}2{\times}2$ tokenizer processes the user-input video, the $1{\times}4{\times}4$ tokenizer processes the top-3 most relevant retrieved videos, and the $1{\times}8{\times}8$ tokenizer processes the remaining ones.
This adaptive tokenization preserves fine-grained details for the most relevant videos while keeping the total token count manageable.

\subsection{Adaptive Token Merging}
\label{sec: method-token-merging}
While retrieval-based dynamic tokenization effectively allocates the token budget, the quadratic complexity of the DiT’s self-attention still leads to high computational cost as the token sequence length increases.
Recent studies \cite{zhang2025faster, xi2025sparse, zhang2025spargeattn} reveal that DiT attention maps are inherently sparse, with only a small subset of entries contributing meaningfully to the output.
Hence, we propose to adaptively merge unresponsive tokens before the attention operation. This strategy avoids redundant computation while preserving the essential context information.

\subsubsection{Frame-level token responsiveness.}
Previous studies~\cite{cai2025mixture, zhang2025vsa, jeong2024dreammotion, yatim2024space} show that the spatially averaged attention features (e.g., query or key features) serve as a strong proxy for measuring how much each frame contributes to the model’s prediction.
Formally, given query, key, and value matrices $Q, K, V \in \mathbb{R}^{N \times D}$ within a DiT block, we represent all tokens belonging to frame $t$ with a single aggregated vector $\bar{K}_t$ obtained by spatially averaging the key features:
\begin{equation}
    \bar{K}_t = \frac{1}{|\mathcal{I}_t|} \sum_{i \in \mathcal{I}_t} K_i,
    \vspace{-1.5mm}
\end{equation}
where $\mathcal{I}_t$ denotes the set of token indices for frame $t$.
Next, we estimate the responsiveness of each frame by computing its maximum attention response to the target queries $Q_{\text{tgt}}$:
\begin{equation}
    R_t = \max_{q \in Q_{\text{tgt}}} (\text{softmax}\!\left(\frac{q \cdot \bar{K}_t^\top}{\sqrt{D}}\right)).
    \vspace{-1.5mm}
\end{equation}
The responsiveness score $R_t$ measures how strongly frame 
$t$ influences the current generation. A higher $R_t$ indicates that at least one target query token strongly attends to that frame whereas low responsive frames can be safely compressed.

\subsubsection{Adaptive token merging.}
We introduce a learnable convolutional operator $\mathcal{C}_\theta$ that adaptively merges tokens belonging to frames with low responsiveness.
Given a set of tokens $\{X_i\}_{i \in \mathcal{I}_t}$ from frame $t$ with low responsiveness score $R_t$, the operator fuses them into a compact representation:
\begin{equation}
    \tilde{X}_t = \mathcal{C}_\theta(\{X_i\}_{i \in \mathcal{I}_t}), 
    \quad \tilde{X}_t \in \mathbb{R}^{\tfrac{N_t}{r} \times D},
    \vspace{-1.5mm}
\end{equation}
where $r$ is the spatial-temporal reduction factor.
We scale $r$ proportionally to the number of conditional videos, as a larger memory context introduces higher redundancy and thus benefits from stronger compression. Empirically, we find that our adaptive merging strategy better preserves generation quality compared to completely discarding low-importance tokens (see Fig.~\ref{fig: discard-merge-comparison}).

\begin{figure}[!t]
    \centering
    \scriptsize
    \includegraphics[width=0.86\columnwidth]
    {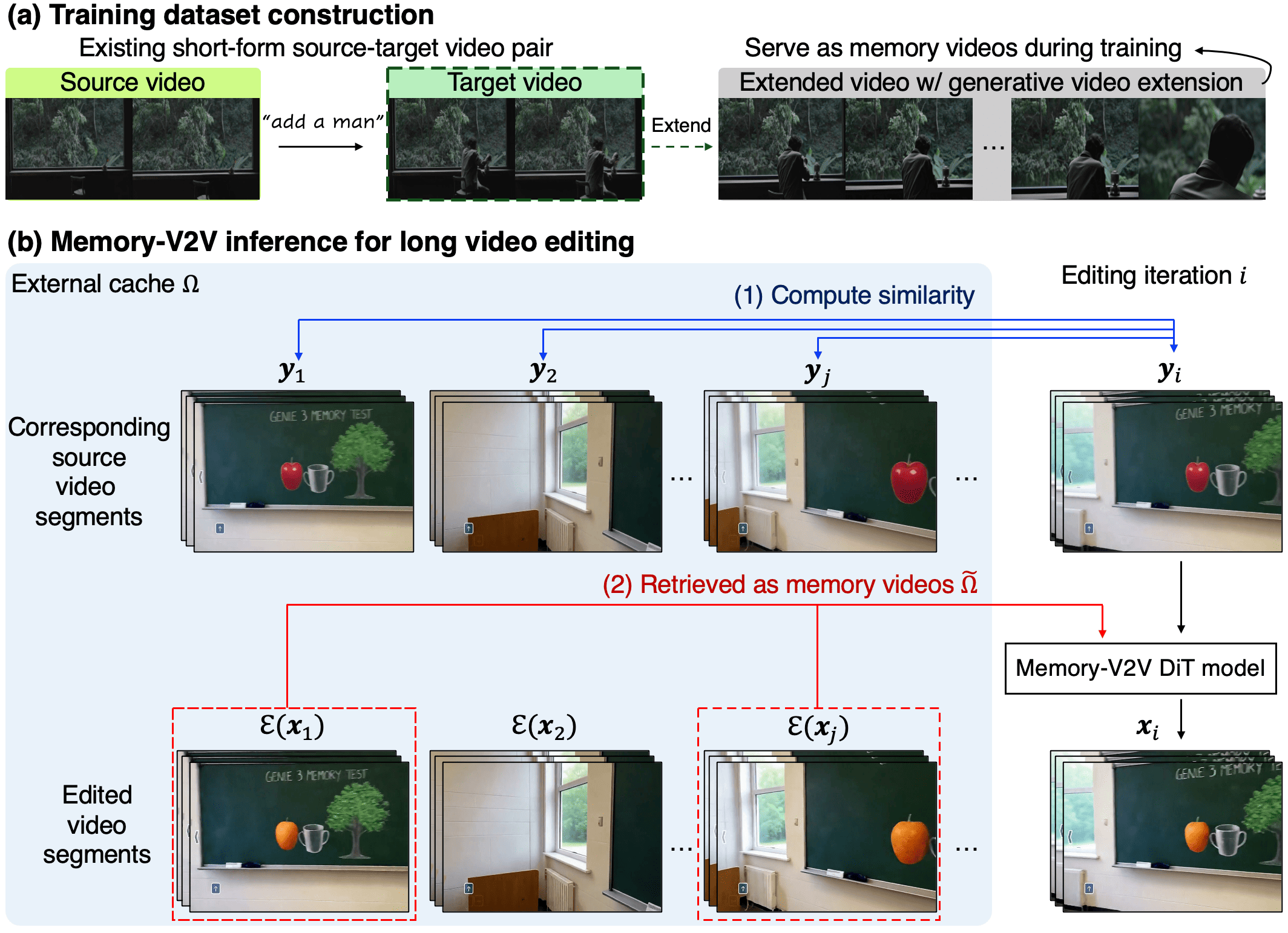}
    \caption{
    \textbf{Long-video editing as multi-turn video editing with Memory-V2V.}
    (a) We extend target videos from an existing video editing dataset for Memory-V2V training.
    (b) During inference, the external cache $\Omega$ stores the editing history as source–target video pairs.
    At the $i$-th editing turn, relevant memory videos are retrieved based on the similarity between source video segments.
    }
    \label{fig: long-video-editing-framework}
\end{figure}

\begin{table}[t]
  \centering
  \scriptsize
  \caption{
  Pearson/Spearman correlations and Bottom-$k$ overlap ($k{=}50\%$) evaluating consistency of frame responsiveness across transformer blocks. 
  Correlations measure linear and rank-order alignment. Bottom-$k$ overlap reflects whether low-responsive frames remain consistently uninformative across layers.
  }
  \begin{adjustbox}{max width=\textwidth}
  \begin{tabular}{lccc}
    \toprule
    &
    {\scriptsize Pearson $r\,\uparrow$} &
    {\scriptsize Spearman $\rho\,\uparrow$} &
    {\scriptsize Bottom-$k$ overlap (\%) $\uparrow$} \\
    \midrule
    DiT Block 1 vs 2--30    & $0.608 \pm 0.137$ & $0.506 \pm 0.267$ & $0.730 \pm 0.160$ \\
    DiT Block 11 vs 12--30  & $0.723 \pm 0.115$ & $0.657 \pm 0.139$ & $0.758 \pm 0.120$ \\
    DiT Block 21 vs 22--30  & $0.753 \pm 0.144$ & $0.683 \pm 0.126$ & $0.793 \pm 0.114$ \\
    \bottomrule
  \end{tabular}
  \end{adjustbox}
  \label{tab:block-corr-bottomk}
\end{table}

\subsubsection{Block selection for token merging.}
To determine where to apply token merging within the DiT architecture, we analyze the stability of responsiveness scores $R_t$ across transformer blocks.
In a 30-block DiT, we group layers into early (1–10), middle (11–20), and late (21–30) stages and measure how well the first block in each stage (1, 11, 21) correlates with subsequent blocks.
As shown in Tab.~\ref{tab:block-corr-bottomk}, the first block (Block 1) shows weak correlation with the rest (Blocks 2–30), meaning early merging risks discarding frames that later become important.
Responsiveness stabilizes in mid and late transformer stages, where low-importance frames remain consistently uninformative. We therefore apply token merging at Blocks 10 and 20, where representations are sufficiently mature for reliable compression.

\begin{figure*}[!t]
    \centering
    \includegraphics[width=\textwidth]{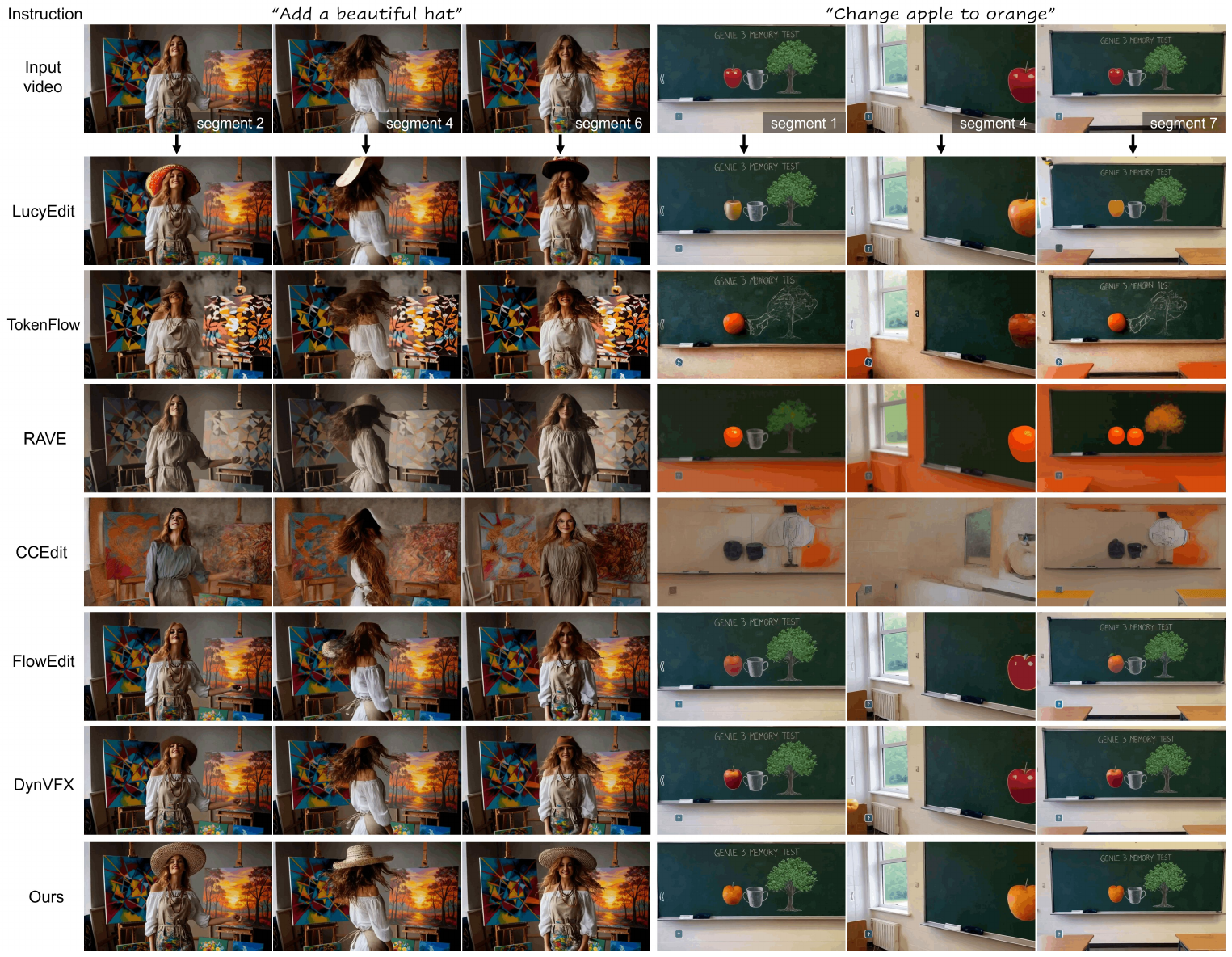}
    \caption{
    \textbf{Text-guided long video editing results by iterative multi-turn editing}.
    In contrast to (iterative) LucyEdit \cite{LucyEdit} or other frame level propagation baselines, Memory-V2V (Ours) consistently inserts the same `hat' (left) and changes apple into the same `orange' (right) across all segments.
    }
    \label{fig: long-video-editing-comparison}
\end{figure*}

\subsection{Extension to Text-guided Long Video Editing}
\label{sec: method-long-video-editing}

We show that Memory-V2V can be applied to other video editing tasks such as text-guided long video editing.
We assume access to a pretrained single-turn text-based video editing model, LucyEdit \cite{LucyEdit}, as our base model in this experiment.
Given a long user-input video $\yb$ typically exceeding 200 frames, which is significantly longer than the temporal context window of the base model (${\sim}$ 81 frames), we reformulate the task as a memory-aware iterative editing. 

We first divide the long source video $\yb$ into shorter segments ${\yb_{1}, \dots, \yb_{T}}$ that fit within the base model’s temporal window and iteratively edit each segment.
When editing the $i$-th source segment $\yb_{i}$, the top-$k$ most relevant edited segments are retrieved from an external cache $\Omega = \{ \mathcal{E}(\xb_{j}) \mid 0 < j < i \}$, which stores all past edited segments as video latents.
Unlike camera-conditioned video novel view synthesis where each generated video $\xb_{j}$ can naturally be indexed by its camera pose $\cb_{j}^{\text{cam}}$ for retrieval, text-conditioned editing poses a unique challenge.
Text instructions are often ambiguous, making text-based similarity unreliable. Instead, we retrieve memory segments based on visual similarity between source video segments using DINO features \cite{oquab2023dinov2}.
Detailed retrieval algorithms are presented in the supplementary.
The retrieved videos are then dynamically tokenized where adaptive token merging is also applied when generating the edited segment $\xb_{i}$.
Once all segments are iteratively edited, they are stitched together to form a single output video $\xb$ (see Fig.~\ref{fig: long-video-editing-framework}(b) for overview).

By casting long video editing as a multi-turn memory-conditioned process, we eliminate the need for long source–target video pairs, which are difficult to obtain in practice.
Instead of requiring long paired data, Memory-V2V can be trained using short source–target pairs by extending only the target videos.
Concretely, starting from a public short video editing dataset, we generatively extend each target video using an off-the-shelf video extension model \cite{zhang2025packing}.
The overall pipeline is visualized in Fig.~\ref{fig: long-video-editing-framework}(a).

\section{Experiments}
\label{sec: experiments}

\begin{table}[!t]
  \centering
  \definecolor{Gray}{gray}{0.9}
  \scriptsize
  \setlength{\tabcolsep}{1.5pt}
  \caption{
  \textbf{Quantitative comparison on multi-turn video novel view synthesis.}
  }
  \label{tab: quantitative nvs comparison}

  \begin{adjustbox}{max width=\columnwidth}
  \begin{tabular}{lcccc|cc|cccc}
    \toprule
      & \multicolumn{4}{c|}{Multi-view Consistency $\downarrow$} & \multicolumn{2}{c|}{Camera Accuracy $\downarrow$} & \multicolumn{4}{c}{Visual Quality $\uparrow$}
      \\
      \cmidrule(lr){2-5} \cmidrule(lr){6-7} \cmidrule(lr){8-11}
      & \makecell{1st Iter. \\ vs 2nd Iter.}
      & \makecell{1st Iter. \\ vs 3rd Iter.}
      & \makecell{2nd Iter. \\ vs 3rd Iter.}
      & \makecell{Avg. Score}
      & \makecell{RotErr}
      & \makecell{TransErr}
      & \makecell{Subject\\Consistency}
      & \makecell{Imaging\\Quality}
      & \makecell{Temporal\\Flickering}
      & \makecell{Motion\\Smoothness}
      \\
    \midrule
    TrajCrafter \cite{yu2025trajectorycrafter}           & 0.1254 & 0.2110 & 0.2090 & 0.1818 & 3.66 & 57.44 & 0.9452 & \textbf{0.7385} & 0.9661 & 0.9880 \\
    ReCam (Ind) \cite{bai2025recammaster}            & 0.1665 & \underline{0.1982} & 0.2031 & 0.1892 & 1.97 & 24.23  & \underline{0.9483} & 0.7186 & \textbf{0.9765} & \underline{0.9931}\\
    ReCam (AR) \cite{bai2025recammaster}             & \underline{0.1181} & 0.1985 & \textbf{0.1290} & \underline{0.1485} & - & - & - & - & - & - \\
    \rowcolor{Gray}
    Memory-V2V (Ours)      & \textbf{0.1168} & \textbf{0.1525} & \underline{0.1379} & \textbf{0.1357} & \textbf{1.65} & \textbf{13.47} & \textbf{0.9494} & \underline{0.7242} & \underline{0.9728} & \textbf{0.9933} \\
    \bottomrule
  \end{tabular}
  \end{adjustbox}
\end{table}

\begin{table}[!t]
  \centering
  \definecolor{Gray}{gray}{0.9}
  \scriptsize
  \setlength{\tabcolsep}{3pt}
  \caption{
  \textbf{Quantitative comparison results for text-guided long video editing.}
  The higher the better for all metrics.
  }
  \begin{adjustbox}{max width=\columnwidth}
  \begin{tabular}{lcccccc|cc}
    \toprule
    
      & \makecell{Subject\\Consistency}
      & \makecell{Background\\Consistency}
      & \makecell{Aesthetic\\Quality}
      & \makecell{Imaging\\Quality}
      & \makecell{Temporal\\Flickering}
      & \makecell{Motion\\Smoothness}
      & \makecell{DINO-F}
      & \makecell{CLIP-F} \\
    \midrule
    LucyEdit (Ind)                 & 0.8683 & 0.9026 & 0.4601 & 0.6429 & 0.9844 & 0.9915 & 0.6856 & 0.8225 \\
    LucyEdit (FIFO)          & 0.8737 & 0.9042 & 0.4527 & 0.5598 & 0.9844 & 0.9919 & 0.6784 & 0.8198 \\
    TokenFlow~\cite{geyer2023tokenflow} & 0.9191 & 0.9172 & 0.4206 & 0.6649 & 0.9845 & 0.9896 & 0.6480 & 0.7563 \\
    RAVE~\cite{kara2024rave} & \underline{0.9220} & \textbf{0.9322} & 0.4703 & 0.6012 & 0.9813 & 0.9872 & 0.7308 & 0.7713 \\
    CCEdit~\cite{feng2023ccedit} & 0.8698 & 0.9101 & 0.4113 & 0.6592 & 0.9788 & 0.9762 & 0.6088 & 0.7228 \\  
    FlowEdit~\cite{kulikov2025flowedit} & 0.9027 & 0.9123 & 0.4865 & 0.6433 & 0.9852 & 0.9931 & \textbf{0.8612} & \textbf{0.8943}   \\
    DynVFX~\cite{yatim2025dynvfxaugmentingrealvideos} & 0.9119 & 0.9210 & \textbf{0.5179} & \underline{0.6666}  & \underline{0.9855} & \underline{0.9937} & \underline{0.8387} & 0.8682\\  
    \rowcolor{Gray}
    Memory-V2V (Ours)         & \textbf{0.9326} & \underline{0.9233} & \underline{0.4950} & \textbf{0.6759} & \textbf{0.9862} & \textbf{0.9939} & 0.8019 & \underline{0.8741} \\
    \bottomrule
  \end{tabular}
  \end{adjustbox}
  \vspace{-1.5mm}
  \label{tab:main_results_vq_feature}
\end{table}

\subsection{Implementation Details}
\label{subsec: implementation details}

\subsubsection{Video novel view synthesis.}
Starting from the single-turn video novel view synthesis model ReCamMaster \cite{bai2025recammaster}, we finetune it for iterative novel view synthesis using a synthetic multi-camera video dataset containing 10 synchronized videos per scene, each captured from distinct camera trajectories.
We finetune the self-attention layers, MLP projector, and camera encoder from the base model, together with the newly introduced dynamic tokenizers and adaptive token compressors.
During training, 1--6 videos are randomly sampled at each iteration to serve as memory videos ${\tilde\Omega}$.
To ensure all tokenizers with varying kernel sizes are trained without reliance on the other, we randomly use one of the tokenizer for the 50\% of the training time, while for the other 50\% we use a mix of them.
Moreover, adaptive token merging is enabled with 50\% probability, using a compression factor $r$ randomly sampled from $[0.3, 0.7]$.

\subsubsection{Text-guided long video editing.}
We extend the single-turn, instruction-driven video editing model LucyEdit \cite{LucyEdit} to a long video editor. We finetune it to take both the source video and previously edited (history) videos as input.
Training is conducted on 56K samples \cite{ju2025editverse} filtered from the publicly available Señorita-2M dataset \cite{zi2025se}.
Each sample in Señorita contains a triplet of text instruction, source video, and target video.
However, since both the source and target clips in Señorita are short-form videos, we employ a generative video extension model \cite{zhang2025packing} to temporally extend the target videos, using these extended segments as the memory videos during training (see Fig.~\ref{fig: long-video-editing-framework}(a)).

Both video novel synthesis and long video editing models are trained with the rectified flow matching loss \cite{esser2024scaling} on 32 A100 GPUs with a total batch size of 32. Due to strong pretrained initialization, stable convergence is achieved within 1–2K finetuning steps.

\begin{table}[!t]
  \centering
  \scriptsize
  \setlength{\tabcolsep}{3pt}
  \caption{
  \textbf{Quantitative ablation on each component of \name.}
  }
  \begin{adjustbox}{max width=\columnwidth}
  \begin{tabular}{lc|c|ccccc}
    \toprule
    
      & \makecell{Time\\(s) $\downarrow$}
      & \makecell{MEt3R $\downarrow$}
      & \makecell{Subject\\Consistency $\uparrow$}
      & \makecell{Aesthetic\\Quality $\uparrow$}
      & \makecell{Imaging\\Quality $\uparrow$}
      & \makecell{Motion\\Smoothness $\uparrow$} \\
    \midrule

    Dynamic Tokenization Only
      & 980.80 & 0.2234 & \textbf{0.9368} & \underline{0.5632} & \textbf{0.7307} & 0.9917 \\
    \cmidrule(lr){1-7}
    + Video Retrieval
      & 965.95 & \textbf{0.2169} & 0.9338 & \underline{0.5632} & 0.7288 & 0.9918 \\
    + Adaptive Token Merging
      & \underline{661.31} & 0.2344 & \underline{0.9361} & 0.5626 & 0.7287 & \textbf{0.9921} \\
    + All (Full Model)
      & \textbf{648.5} & \underline{0.2208} & 0.9351 & \textbf{0.5636} & \underline{0.7300} & \underline{0.9919} \\
    \bottomrule
  \end{tabular}
  \end{adjustbox}
  \label{tab:ablation_dynamic_tokenization}
\end{table}

\subsection{Memory-V2V for Video Novel View Synthesis}
\label{subsec: experiments vnv}

In this experiment, we evaluate on 40 publicly available videos \cite{nan2024openvid, bai2025recammaster, yu2025trajectorycrafter, wiedemer2025video}, each serving as a user input.
We compare \name against ReCamMaster (ReCam) \cite{bai2025recammaster} and TrajectoryCrafter (TrajCrafter) \cite{yu2025trajectorycrafter}.
For ReCamMaster, two inference modes are evaluated:
(1) ReCam (Ind), which reuses the same user-input video as the source for every iteration, and
(2) ReCam (AR), which uses the previous output as the next input in a sequential manner.
ReCam (Ind) avoids error accumulation but lacks cross-turn memory, while ReCam (AR) enforces sequential conditioning yet accumulates drift across iterations.
Qualitative comparisons are shown in Fig.~\ref{fig: novel-view-synthesis-comparison}, and quantitative results in Tab.~\ref{tab: quantitative nvs comparison}.
For each input video, we iteratively generate three outputs with highly overlapping camera trajectories (e.g., pan-left and orbit-right) and measure pairwise consistency across the 1st–2nd, 1st–3rd, and 2nd–3rd iterations.
We adopt MEt3R to measure cross-view geometric consistency, VBench \cite{huang2024vbench} to evaluate visual quality, and report rotation \& translation errors. We observe that as the number of iterations increases, the cross-consistency significantly drops for the base models (TrajCrafter and ReCam (Ind)). While ReCam (AR) preserves consistency between subsequent iterations (e.g., 2nd and 3rd), it fails to keep other iterations (e.g., 1st and 3rd) consistent. In contrast, \name results in consistent generations across all iterations while successfully preserving the visual quality and improving camera adherence of the base model.

\subsection{Memory-V2V for Long Video Editing}
\label{subsec: experiments videdit}

We evaluate the long video editing performance on 50 videos from the Señorita test set \cite{zi2025se}.
We first compare against two variants of LucyEdit \cite{LucyEdit}: LucyEdit (Ind), which edits each video segment independently, and LucyEdit (FIFO), which follows FIFO-Diffusion’s diagonal denoising strategy to process consecutive frames with increasing noise levels, simulating autoregressive generation \cite{kim2024fifo}.In addition, we consider five more baselines: TokenFlow \cite{geyer2023tokenflow}, which propagates diffusion features across frames to encourage temporal consistency; RAVE \cite{kara2024rave}, which employs randomized noise shuffling to produce consistent videos while enabling fast video editing; CCEdit \cite{feng2023ccedit}, which uses keyframe appearance and structure guidance for consistent editing; FlowEdit \cite{kulikov2025flowedit} performs inversion-free, optimization-free prompt based editing via flow matching; and DynVFX \cite{yatim2025dynvfxaugmentingrealvideos} enables zero-shot video augmentation by manipulating attention features in a pretrained T2V model.
As shown in Fig.~\ref{fig: long-video-editing-comparison}, while TokenFlow and RAVE can maintain coarse edited-object consistency across frames, they often fail to preserve fine-grained subject identity and unintentionally modify background regions during editing. 
CCEdit frequently exhibits limited editing controllability, leading to incomplete or inconsistent edits. 
LucyEdit preserves subject identity and background structure via token reuse through self-attention, but still struggles to maintain edited-object consistency over long sequences. FlowEdit often fails to follow the instruction, leaving the target object unedited or inconsistently edited across frames. DynVFX can edit the object, but may generate the new object at unintended locations rather than replacing the original one as shown in Fig.~\ref{fig: long-video-editing-comparison}. In contrast, Memory-V2V simultaneously preserves subject identity, background structure, and edited-object consistency by leveraging explicit memory conditioning, enabling stable and coherent edits across extended video sequences (${>}200$ frames).
We report visual quality and consistency metrics in Tab.~\ref{tab:main_results_vq_feature}, using VBench and cross-frame DINO/CLIP similarity metrics \cite{oquab2023dinov2, radford2021learning}, where Memory-V2V consistently outperforms all baselines. Cross-frame DINO/CLIP similarity should be considered alongside edit success: overly high values may indicate insufficient editing, whereas overly low values can suggest unintended changes or temporal instability.

\begin{figure}[t]
    \centering
    \scriptsize
    \includegraphics[width=0.80\columnwidth]{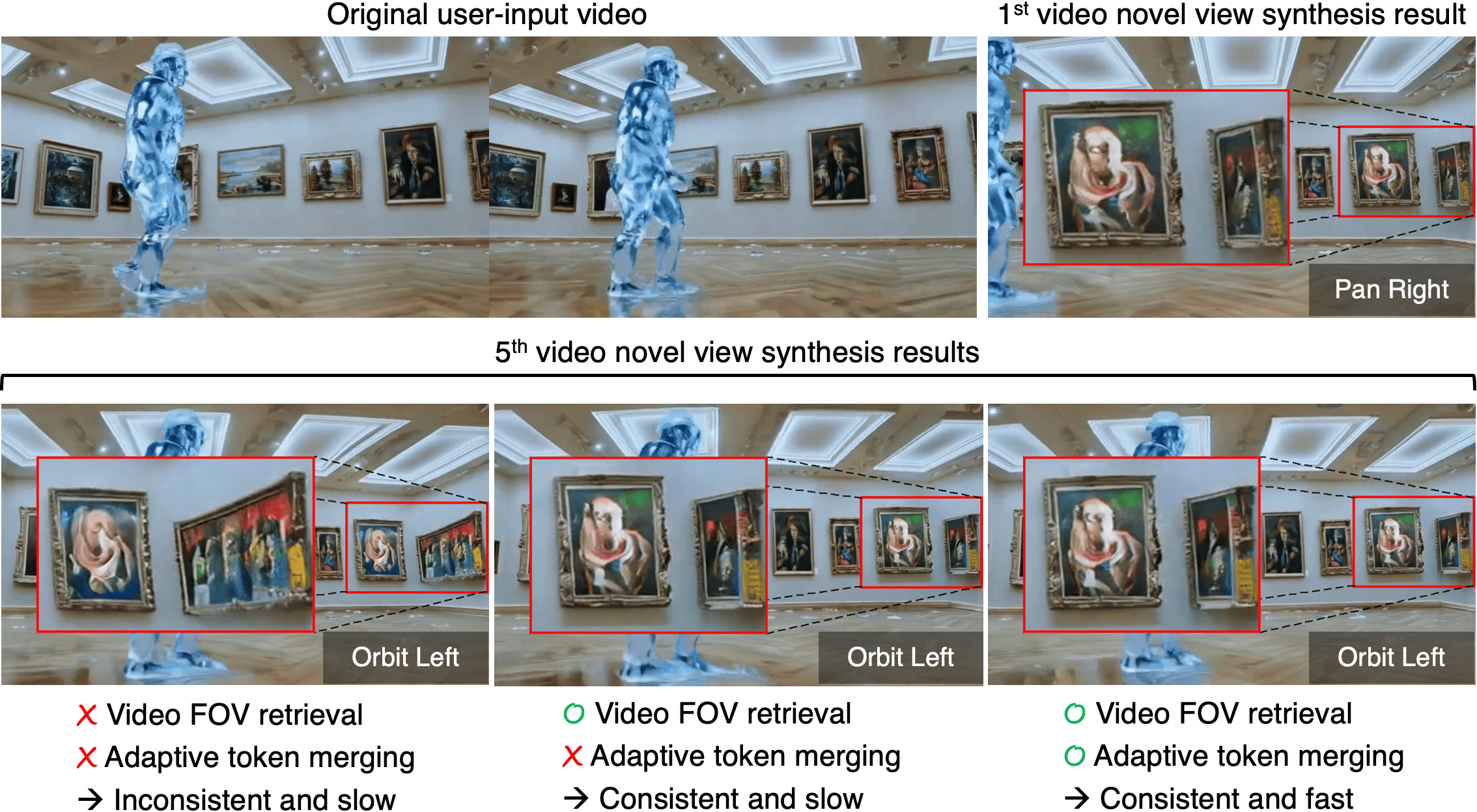}
    \caption{\textbf{Qualitative ablation on video retrieval and adaptive token merging.}}
    \label{fig: nvs-ablation}
    \vspace{-5mm}
\end{figure}

\begin{figure}[t]
    \scriptsize
    \centering
    \begin{minipage}{0.71\columnwidth}
        \centering
        \includegraphics[width=\linewidth]{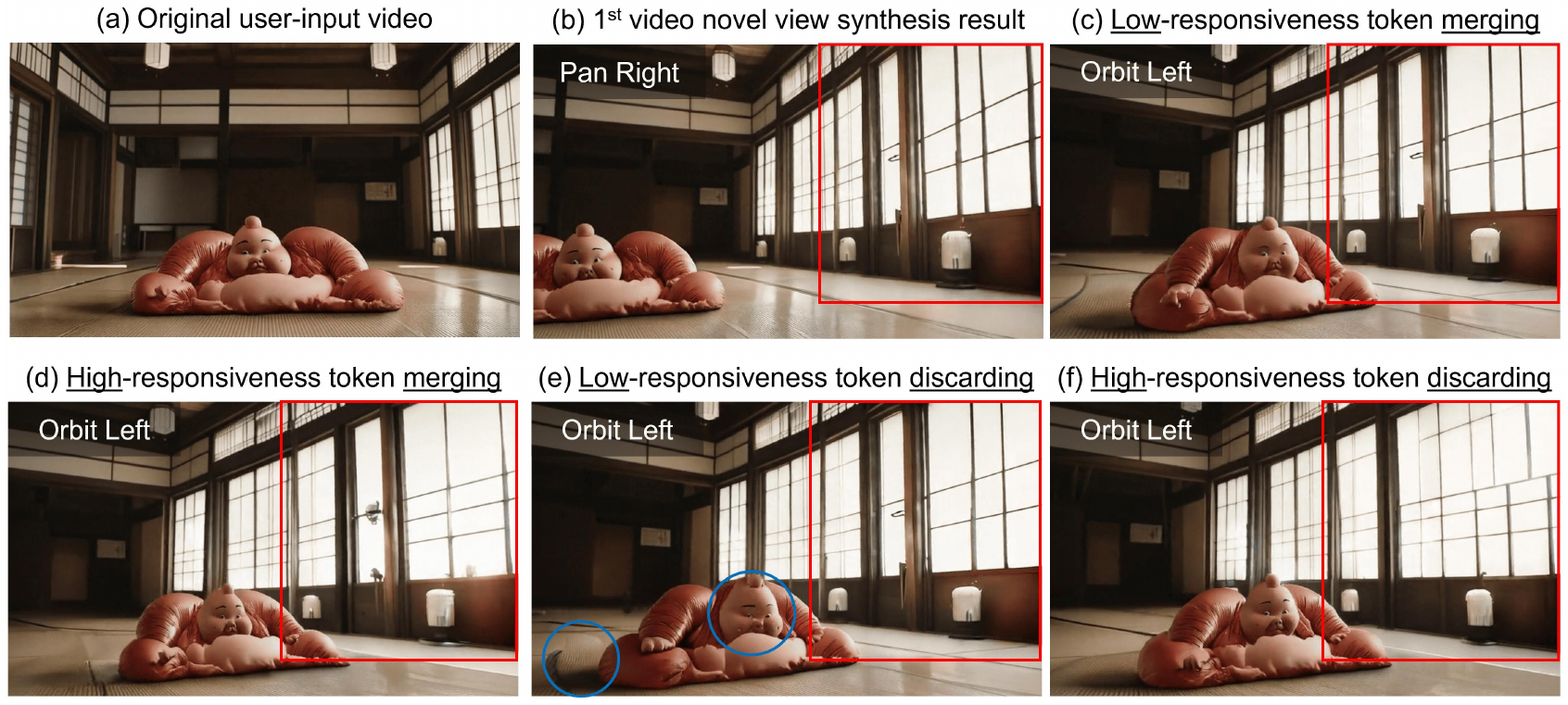}
        \vspace{-13mm}
        \caption{
        \textbf{Qualitative ablation on token responsiveness and reduction strategies.}
        }
        \label{fig: discard-merge-comparison}
    \end{minipage}
    \hfill
    \begin{minipage}{0.28\columnwidth}
        \centering
        \includegraphics[width=\linewidth]{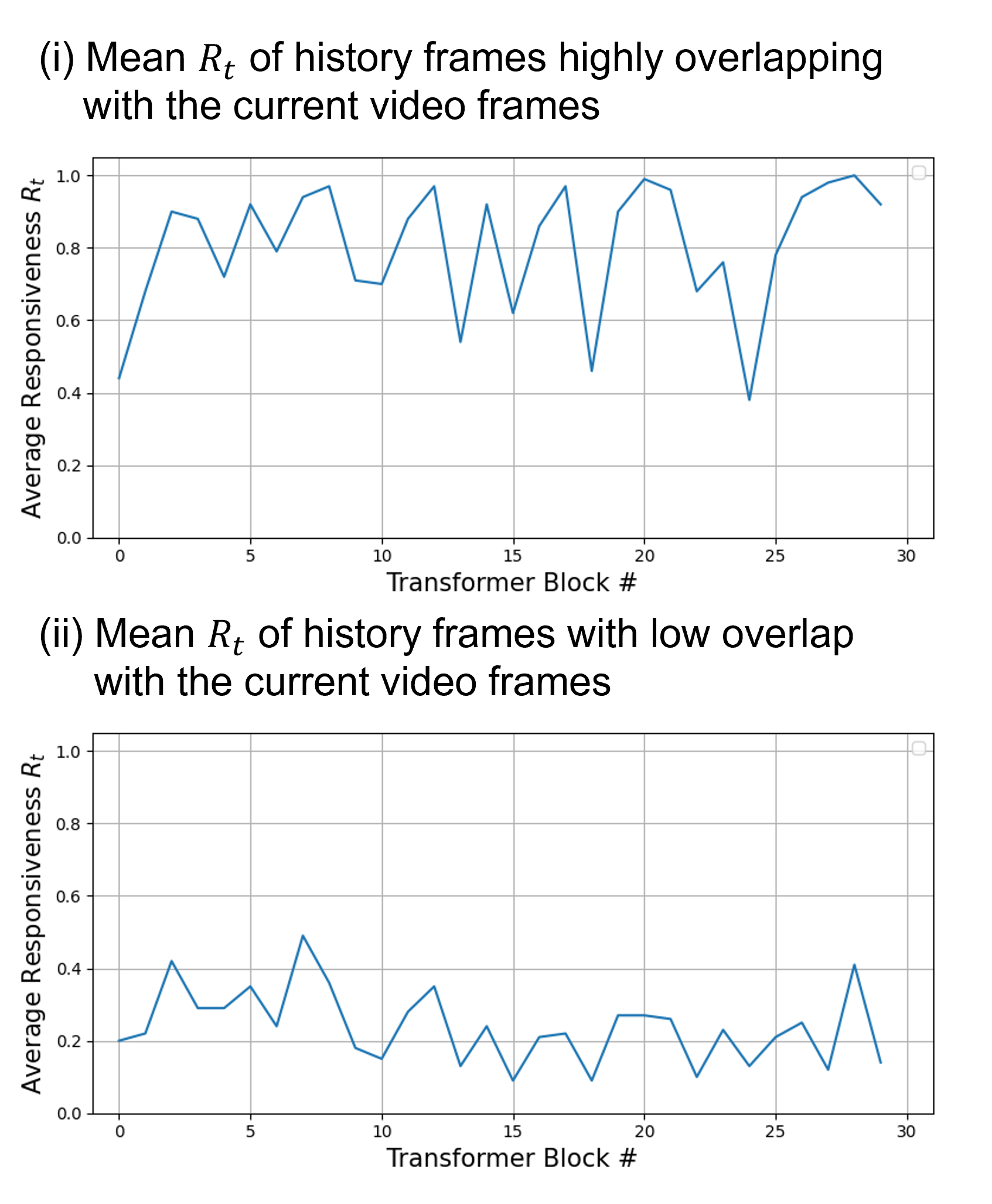}
        \caption{\textbf{Mean responsiveness $R_t$ analysis.}}
        \label{fig: token-analysis}
    \end{minipage}
    \vspace{-5mm}
\end{figure}

\subsection{Ablation Study}

\subsubsection{Ablation on each component.}
We first ablate the effects of main components of \name, as summarized in Tab.~\ref{tab:ablation_dynamic_tokenization} and Fig.~\ref{fig: nvs-ablation}.
Dynamic tokenization combined with video retrieval improves cross-iteration consistency, while adaptive token merging notably reduces computational cost.
As shown in Fig.~\ref{fig: nvs-ablation}, retrieval guided by VideoFOV effectively preserves long-term consistency even after multiple iterations (e.g., between the 1st and 5th generations), whereas token merging maintains efficiency without visible degradation.

\subsubsection{Ablation on token responsiveness and token reduction.}
We analyze the effect of conditioning tokens ranked by their responsiveness (Sec.~\ref{sec: method-token-merging}) under two reduction strategies: merging and discarding.
\footnote{Tokens are partitioned by responsiveness score $R_t$: the top 50\% are high-responsive, and the bottom 50\% are low-responsive. Additional experimental details are provided in the supplementary material.}
Multi-turn video editing needs to achieve cross-iteration consistency across edits while preserving the visual quality of the generated videos.

Highly responsive conditioning tokens are essential for cross-consistency across edits.
When these tokens are either merged (Fig.~\ref{fig: discard-merge-comparison}(d)) or discarded (f), the overlapping novel-view regions become inconsistent. The red-boxed region, which should match (b), fails to remain consistent in both cases.
This observation is also supported by the frame-responsiveness analysis in Fig.~\ref{fig: token-analysis}(i), where high-responsive tokens are strongly correlated with the target camera trajectory.

Discarding low-responsive tokens may degrade fine visual details and motion.
Fig.~\ref{fig: discard-merge-comparison} shows that discarding these tokens introduces unnatural artifacts in the blue-highlighted regions. 
Furthermore, Fig.~\ref{fig: token-analysis}(ii) shows that these tokens are less correlated with the video frames generated in the current turn,
but they may still contribute to structural consistency.
Therefore, instead of discarding them,
merging low-responsive tokens reduces computation without affecting cross-consistency and visual fidelity. 
Adaptive merging preserves cross-turn consistency while reducing computation, whereas discarding low-responsive tokens introduces visual artifacts. Additional examples are provided on the supplementary project page.

\vspace{-2mm}
\subsubsection{Computational cost analysis.}
\begin{figure}[t]
    \centering
    \includegraphics[width=\columnwidth]
    {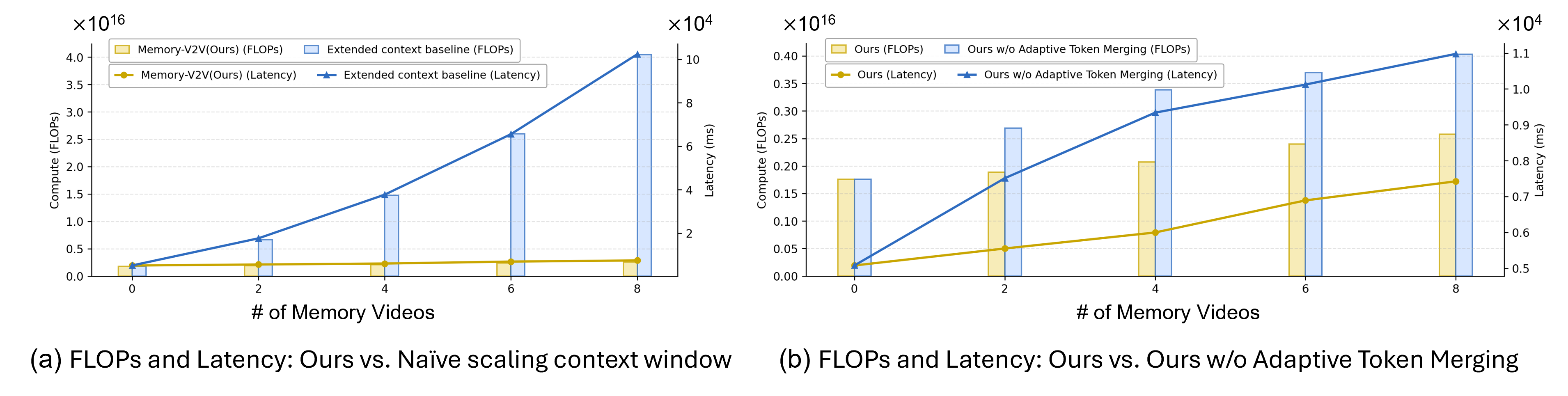}
    \caption{
    \textbf{Computational cost (FLOPs, Latency) analysis.} 
    (a) Comparison between \name and naive context-window scaling for memory videos.
    (b) Comparison of \name with and without adaptive token merging.
    Notably, the computational gains increase as the number of memory videos grows.
    }
    \label{fig: FLOPs-graph}
    \vspace{-5mm}

\end{figure}

We analyze the computational gains of \name in Fig.~\ref{fig: FLOPs-graph}. 
As shown in Fig.~\ref{fig: FLOPs-graph}(a), compared to naive context-window scaling, our method reduces FLOPs and latency by over 90\%.
Furthermore, Fig.~\ref{fig: FLOPs-graph}(b) shows that adaptive token merging further reduces FLOPs and latency by 30\%.

\section{Conclusion}
\label{sec: conclusion}

We formulated multi-turn video editing as a distinct consistency challenge and introduced \name, a memory-augmented framework that treats prior edits as structured constraints. By combining task-aware retrieval and relevance-aware compression, \name enables scalable and consistent iterative editing across diverse video tasks.
\vspace{-3mm}
\subsubsection{Limitations.} \name inherits the limitations (e.g., large view changes) of the underlying single-turn editing model on which it is built. Additionally, since the training data consists only of continuous single-shot videos, \name may struggle with editing multi-shot long videos containing abrupt scene transitions. Multi-shot video editing is an exciting research direction we would like to explore in the future.
\vspace{-2mm}

\section*{Acknowledgements}
\vspace{-2mm}
We thank Minguk Kang and Sihyun Yu for their insightful discussions at the early stage of the project. This work was supported by the National Research Foundation of Korea under Grant RS-2024-00336454. This work was supported by the Institute of Information \& communications Technology Planning \& Evaluation (IITP) grant funded by the Korean government(MSIT) (No. RS-2024-00457882, AI Research Hub Project). This research was supported by the Advanced GPU Utilization Support Program funded by the Government of the Republic of Korea (Ministry of Science and ICT) (02-26-01-0404).

\clearpage
\begin{appendix}

\def\vx{{\boldsymbol{x}}}
\def\x{{\boldsymbol{x}}}

This supplementary material is organized as follows.
Sec.~\ref{sec: flow_matching} introduces the rectified flow matching loss used to train Memory-V2V.
Sec.~\ref{sec: details_on_ablation} provides experimental details for the ablation study.
Sec.~\ref{sec: method-toy-experiment} presents toy experiments analyzing suitable context encoders for memory representation, and Sec.~\ref{sec: ideal_context_encoder_detail} provides further implementation details and results.
Sec.~\ref{sec: video_retrieval} describes the retrieval mechanisms for video novel view synthesis and text-guided long video editing.
Sec.~\ref{sec: multi_view_consistency} evaluates multi-view consistency against ground-truth target-view videos using synchronized multi-camera scenes.
Sec.~\ref{sec: training_inference_detail} details the training and inference setups, including RoPE design and camera conditioning.
Sec.~\ref{sec: recent_v2v} compares Memory-V2V with a recent video-to-video model, highlighting its improved global consistency over long, segment-wise edited videos.
Sec.~\ref{sec: object_level_consistency} demonstrates that Memory-V2V preserves the identity and 3D location of edited objects across revisited viewpoints and multiple editing turns.
Sec.~\ref{sec: additional_qual} presents additional qualitative results.
Finally, Sec.~\ref{sec: discussion} discusses limitations, failure cases, and future directions.
A full set of video results is available on the project page.

\section{Preliminary: Rectified Flow Matching}
\label{sec: flow_matching}

Recently, flow-based training frameworks~\cite{liu2022flow, esser2024scalingrectifiedflowtransformers, chen2025goku, opensora} have gained attention for their ability to progressively transform samples from a prior distribution to the target data distribution through a series of linear interpolations. 
Specifically, we define a velocity field $v_t(\vx)$ of a flow $\psi_t(\vx): [0,1]\times \mathbb{R}^d \rightarrow \mathbb{R}^d$ that satisfies $\psi_t(\x_0)=\x_t$ and  $\psi_1(\x_0)=x_1$.
Here, the $\psi_t$ is uniquely characterized by a flow ODE:
\begin{align}
    d\psi_t(\vx) = v_t(\psi_t(\vx))dt
    \label{eqn:flowode}
\end{align}

In particular, linear conditional flow defines the flow as $\x_t=\psi_t(\vx_1|\vx_0) = (1-t)\vx_0 + t \vx_1$. Then, we can compute the velocity field $v_t(\vx_t|\vx_0) = \dot \psi_t(\psi_t^{-1}(\vx_t|\vx_0)|\vx_0)=\vx_1-\vx_0$, leading to the
following conditional flow matching loss:

\begin{equation}
\label{eq: flow loss}
L_{\mathrm{RF}} = \mathbb{E}_{t, \, \vx_0,\, \boldsymbol{\epsilon} \sim \mathcal{N}(\mathbf{0},\mathbf{I})} \left[ \left\| (\vx_0 - \boldsymbol{\epsilon}) - v_\theta(\vx_t, t) \right\|^2 \right].
\end{equation}
During inference, the predicted velocity is utilized to guide the transformation of an initial noise sample towards the target data distribution through a reverse integration process. This approach enables efficient and controlled generation of high-quality samples.
\clearpage

\section{Additional Ablation Study Details}
\label{sec: details_on_ablation}
\vspace{-2mm}

\subsection{Video retrieval and adaptive token merging experiments.}
We design this experiment to analyze how video retrieval and adaptive token merging affect long-term consistency.
Specifically, the first and fifth iterations use camera trajectories with overlapping visible regions (e.g., pan-right followed by orbit-left, or pan-left followed by orbit-right), while the remaining iterations use randomly sampled trajectories.

For runtime and visual quality evaluation, metrics are computed on the fifth-iteration outputs. For MEt3R, we measure consistency between the videos generated in the first and fifth iterations to assess whether multi-view consistency is preserved over long editing sequences.
The results show that video retrieval improves cross-iteration multi-view consistency, while adaptive token merging significantly reduces runtime without degrading generation quality or consistency.

\vspace{-3mm}
\subsection{Token reduction experiments.}
In Fig.~7 (main) and Fig.~8 (main), we conduct three-turn novel view synthesis.
The first and third iterations use camera trajectories with overlapping viewpoints, while the second iteration uses an unrelated trajectory.
For example, when the first and third iterations correspond to pan-right and orbit-left, the second iteration uses a trajectory such as tilt-up with minimal overlap.

During the third iteration, we manipulate tokens from 50\% of the frames by either merging or discarding them.
For high-responsiveness manipulation, we operate on tokens from the top 50\% of frames with the largest responsiveness scores ($R_t$).
For low-responsiveness manipulation, we operate on tokens from the bottom 50\% of frames with the smallest responsiveness scores.

For Fig.~9, we analyze the responsiveness of historical frames when generating the third iteration. We first measure the responsiveness of frames that are geometrically related to the target camera trajectory (e.g., frames from videos with trajectories such as pan-left when generating orbit-right). The responsiveness scores are averaged across transformer blocks.

We then perform the same analysis for frames unrelated to the target trajectory, including frames from the static source video and from videos generated with unrelated trajectories (e.g., tilt-up). Their responsiveness scores are also averaged across blocks.

As shown in Fig.~9 (main), frames with high responsiveness in early transformer blocks tend to remain consistently high across later blocks, while low-responsive frames remain consistently low. Moreover, frames geometrically aligned with the target trajectory generally exhibit higher responsiveness than unrelated frames.

\section{Proof-of-Concept Experiment: Ideal Context Encoder}
\label{sec: method-toy-experiment}
\vspace{-1mm}
Maintaining semantic consistency and persistency when editing long videos in chunks or in multi-turn editing setup requires to \emph{encode} prior editing results as additional context which can then be utilized as constraints for the next editing round. Hence, an important design choice is how the context is represented. We devise a simple experiment setup to evaluate the representation capability of different potential context encodings. We limit our experiment to the novel-view synthesis task in a constrained two-turn editing setup using ReCamMaster as the base model. In particular, given an input video $\yb$, in the first turn we generate a video $\xb_{1}$ from a new camera trajectory $\cb_{1}^\text{cam}$.
Using an arbitrary encoder $\mathcal{E}$, we extract the context $\mathcal{E}(\xb_{1})$ from $\xb_{1}$ and in the second turn, we generate a video $\xb_{2}$ from a camera trajectory $\cb_{2}^\text{cam}$, conditioning on both the original source video $\yb$ as well as the encoded context $\mathcal{E}(\xb_{1})$. If $\mathcal{E}(\xb_{1})$ has sufficient representation power, we expect that the novel content generated in $\xb_{1}$ and $\xb_{2}$ to be consistent.

\begin{figure}[!t]
    \centering
    \includegraphics[width=0.9\columnwidth]
    {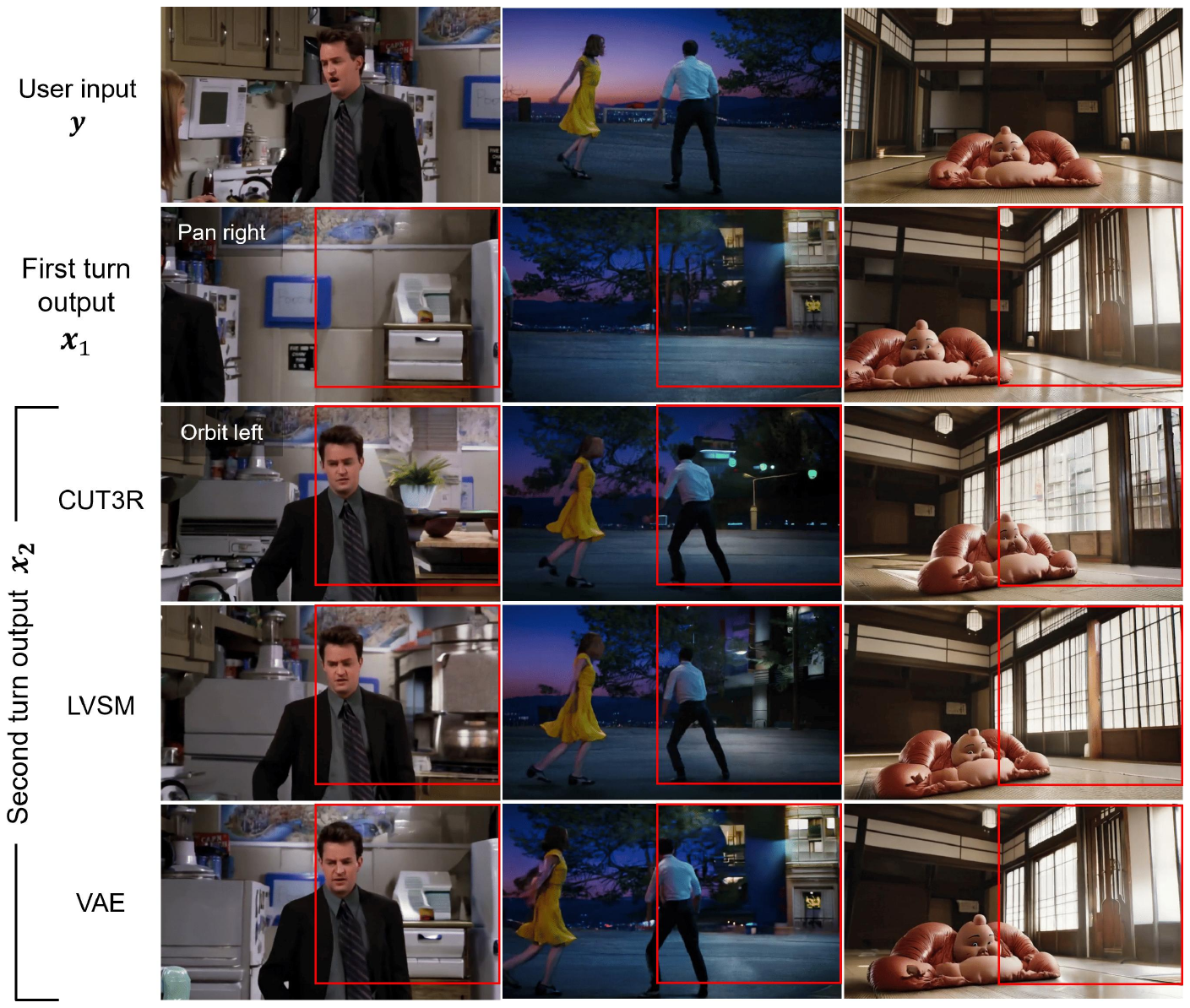}
    \caption{
    \textbf{Comparison of different memory encoders on two-turn novel view synthesis.} 
    The red-colored box depicts the novel region which are expected to be consistent between $\xb_{1}$ and $\xb_{2}$.
    }
    \vspace{-3mm}
    \label{fig: ideal-memory-representation}
\end{figure}

We experiment with three different choices of pretrained context encoders for the novel view synthesis task: 
(i) a recurrent 3D reconstructor model CUT3R~\cite{wang2025continuous}
(ii) a state of the art novel view synthesis network LVSM~\cite{jin2024lvsm}, and
(iii) a video VAE~\cite{kingma2013auto}.
For each setting, encoded representations $\mathcal{E}(\xb_{1})$ are patchified to match the dimensionality of the DiT's video tokens, allowing interaction within the self-attention layers.
During finetuning, we freeze the context encoder $\mathcal{E}$ and train only the patchification layer and the ReCamMaster model to accept additional memory context (see the supplementary for details).
Moreover, during training, we randomly sample multiple videos of the same scene: one as the original source video $\yb$ and another as the source of contextual memory $\xb_{1}$.

We provide examples from each of the context encodings in Fig.~\ref{fig: ideal-memory-representation}. We observe that the state tokens from CUT3R or LVSM fall short for encoding detailed appearance in newly generated regions. In contrary, using the same video latent space achieves the best performance in terms of quality and consistency of the results. Hence, we adopt this option for the context representation and focus on how to efficiently retrieve and process the context representations obtained from multiple past generations as will be discussed next.

\subsection{Context Encoder Setup}
\label{sec: ideal_context_encoder_detail}

\subsubsection{Recurrent 3D reconstructor CUT3R.} 
CUT3R~\cite{wang2025continuous} is designed to predict the 3D point representation rom a video stream by maintaining a recurrent geometric state.
Given an input frame $I_t$, the model first extracts a feature map using a ViT encoder,
\[
F_t = \mathrm{Encoder}(I_t).
\]
To aggregate geometry over time, these features are passed to a decoder along with the previous state $s_{t-1}$, producing
\[
F_t',\; s_t = \mathrm{Decoder}(F_t, s_{t-1}),
\]
where the updated state $s_t \in \mathbb{R}^{768 \times 768}$ accumulates 3D cues across frames that function as the model's internal geometric memory.
3D point map is then predicted from $F_t'$.

To evaluate whether this recurrent state can serve as a memory representation, we sequentially update $s_t$ over video frames, project the state via a learnable MLP into the transformer latent space, and condition the diffusion model through a dedicated branch (following the multimodal conditioning strategy used in SD3~\cite{esser2024scaling}).
The branch shares self-attention with the backbone while using separate modulation and feed-forward layers.
In specific, we fine-tune for 2K steps with a batch size of 32.
However, the diffusion transformer backbone fails to effectively leverage the CUT3R state: generations remain unstable and lack coherent geometry, as can be seen in Fig.~\ref{fig: ideal-memory-representation}.

\begin{figure}[!t]
\centering

\begin{minipage}[t]{0.60\columnwidth}
    \centering
    \includegraphics[width=\linewidth]{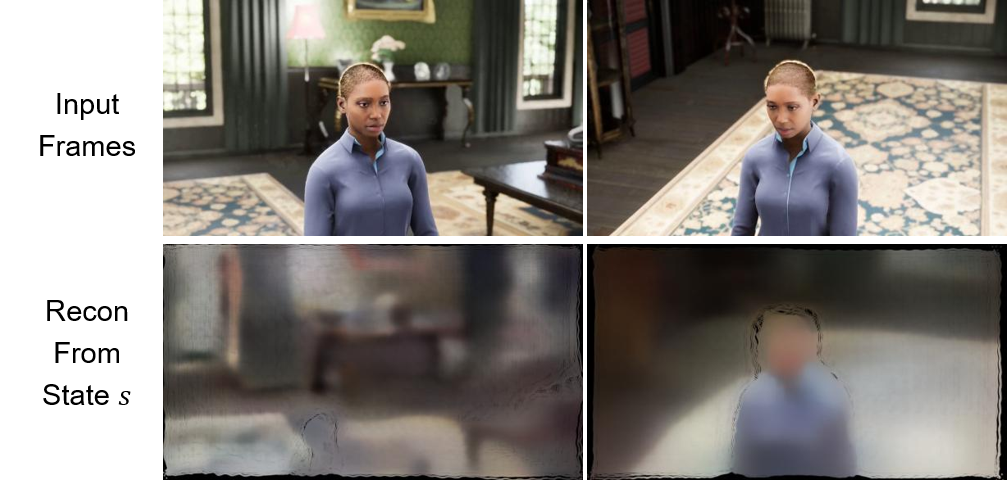}
    \caption{\textbf{Reconstruction results using CUT3R state $s$.}}
    \label{fig: cut3r-recon}
\end{minipage}
\hfill
\begin{minipage}[t]{0.39\columnwidth}
    \centering
    \includegraphics[width=\linewidth]{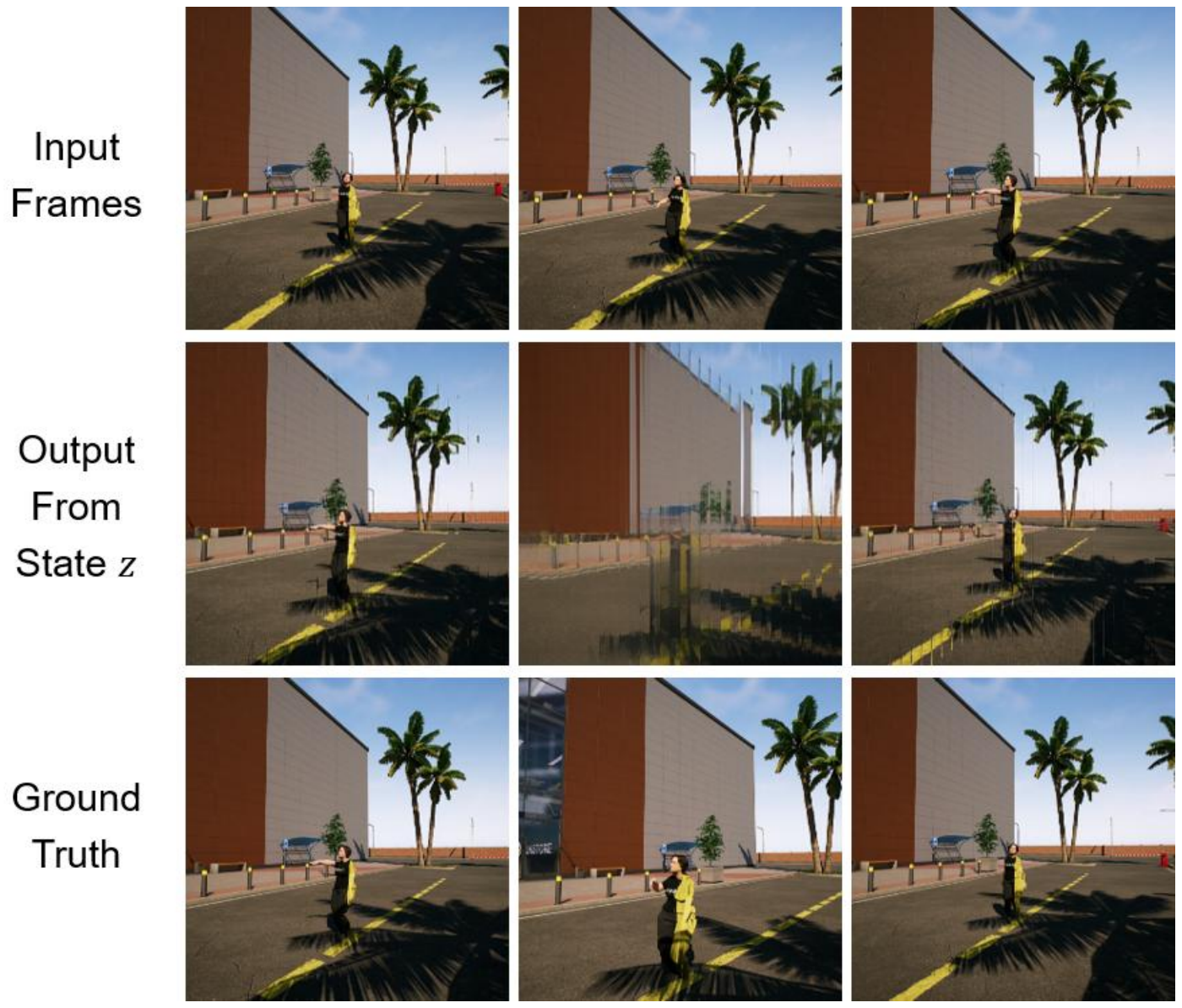}
    \caption{\textbf{Novel view synthesis results using LVSM state $z$.}}
    \label{fig: lvsm-nvs}
\end{minipage}

\end{figure}

These observations further align with our reconstruction analysis.
Following CUT3R’s original procedure, we attempted to reconstruct scenes using only the recurrent state and a ray map.
As shown in Fig.~\ref{fig: cut3r-recon}, the resulting geometry is coarse and blurry, lacking meaningful fine-grained detail.
This suggests that the recurrent state encodes limited and lossy structural information.
As a result, when used as a conditioning signal, it fails to provide sufficiently informative guidance for generation—leading to the degraded performance observed in our CUT3R-conditioned model.

\subsubsection{3D Novel view synthesis model LVSM.} 
LVSM~\cite{jin2024lvsm} synthesizes novel views by jointly encoding input images and their corresponding Plücker ray embeddings.
Given an input image $I_i$ and Plücker embedding $\tilde{P}_i$, the model projects their concatenation into a shared feature space:
\[
x_i = \mathrm{Linear}([I_i,\, \tilde{P}_i]) \in \mathbb{R}^{d}.
\]
These tokens, together with an initial latent token $e$, are processed by the encoder,
\[
x_1, x_2, \dots, x_n,\, z = \mathrm{Encoder}(x_1, x_2, \dots, x_n, e),
\]
producing a latent state $z$. 
For novel view rendering, the decoder conditions on the Plücker embedding $q$ of the target camera rays and predicts
\[
z',\, y = \mathrm{Decoder}(z, q),
\]
where $y$ is the rendered view.

We evaluate whether the latent state $z$ could serve as a meaningful memory representation.
Similar to the CUT3R setup, we project $z$ into the DiT feature dimension via a learnable MLP and inject it through a dedicated conditioning branch that shares self-attention layers with the main model but uses separate modulation and feed-forward layers.
The model is trained for 2K steps with batch size 32.

However, as illustrated in Fig.~\ref{fig: ideal-memory-representation}, conditioning on $z$ provides little benefit.
While LVSM’s decoder can exploit $z$ for its own renderer, the representation lacks the transferable, fine-grained geometric cues needed to guide a diffusion-based generator.
Consequently, the conditioned synthesis is unstable and does not improve multi-turn consistency, suggesting that the LVSM latent space is not structured for effective memory conditioning.

\begin{figure}[!t]
    \centering
    \includegraphics[width=0.7\columnwidth]
    {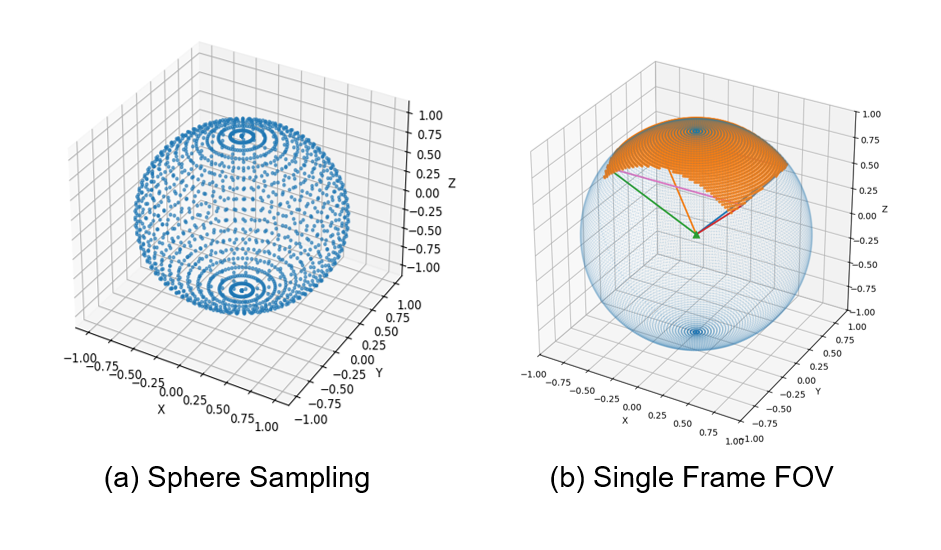}
    \caption{
    \textbf{Sphere sampling and Field-of-View visualization.} 
    }
    \label{fig: sphere_visualization}
\end{figure}

\section{Video Retrieval Algorithms}
\label{sec: video_retrieval}
\subsection{Retrieval for Video Novel View Synthesis}

To compute the VideoFOV retrieval score, we first uniformly sample a dense set of points on a unit sphere centered at the target camera position (see Fig.~\ref{fig: sphere_visualization}(a)).
In specific, we generate a latitude–longitude grid with:
\[
N_\theta = 180, \qquad N_\phi = 360,
\]
resulting in a total of $|M| = N_\theta N_\phi = 64{,}800$ sampled directions. 
The sampling index set is defined as:
\begin{equation}
M
=
\left\{
(u,v)
\;\middle|\;
u \in \{0,\dots,N_\theta{-}1\},\;
v \in \{0,\dots,N_\phi{-}1\}
\right\},
\end{equation}
where each pair $(u,v)$ uniquely maps to a 3D point on the sphere.

Fig.~\ref{fig: sphere_visualization}(b) illustrates how the sampled spherical points are tested for visibility given camera intrinsics and extrinsics. 
During training, intrinsics are taken directly from the dataset; during inference, we adopt a fixed intrinsic matrix from the training set. 
To ensure consistent evaluation, each frame’s camera pose is represented as a relative transform with respect to the first frame of the target video.

Each sampled point $(u,v)\in M$ is then projected to the image plane.
point is considered visible if it falls within image bounds with a valid (positive) depth value.
The per-frame field of view (FOV) is defined as:
\begin{equation}
\mathcal{F}_\text{frame}(\cb_{i,t}^\text{cam})
=
\left\{
(u,v) \in M
\;\middle|\;
(u,v)\ \text{projects inside the image} 
\right\}.
\end{equation}
We accumulate these visibility sets across all frames to compute the video-level FOV, and then evaluate the relevance score using Eq. (1), (2), and (3). The complete retrieval workflow is provided in Alg.~\ref{alg:video-fov-retrieval}.

\begin{figure}[!t]
\centering

\begin{minipage}[t]{0.49\columnwidth}
    \centering
    \includegraphics[width=\linewidth]{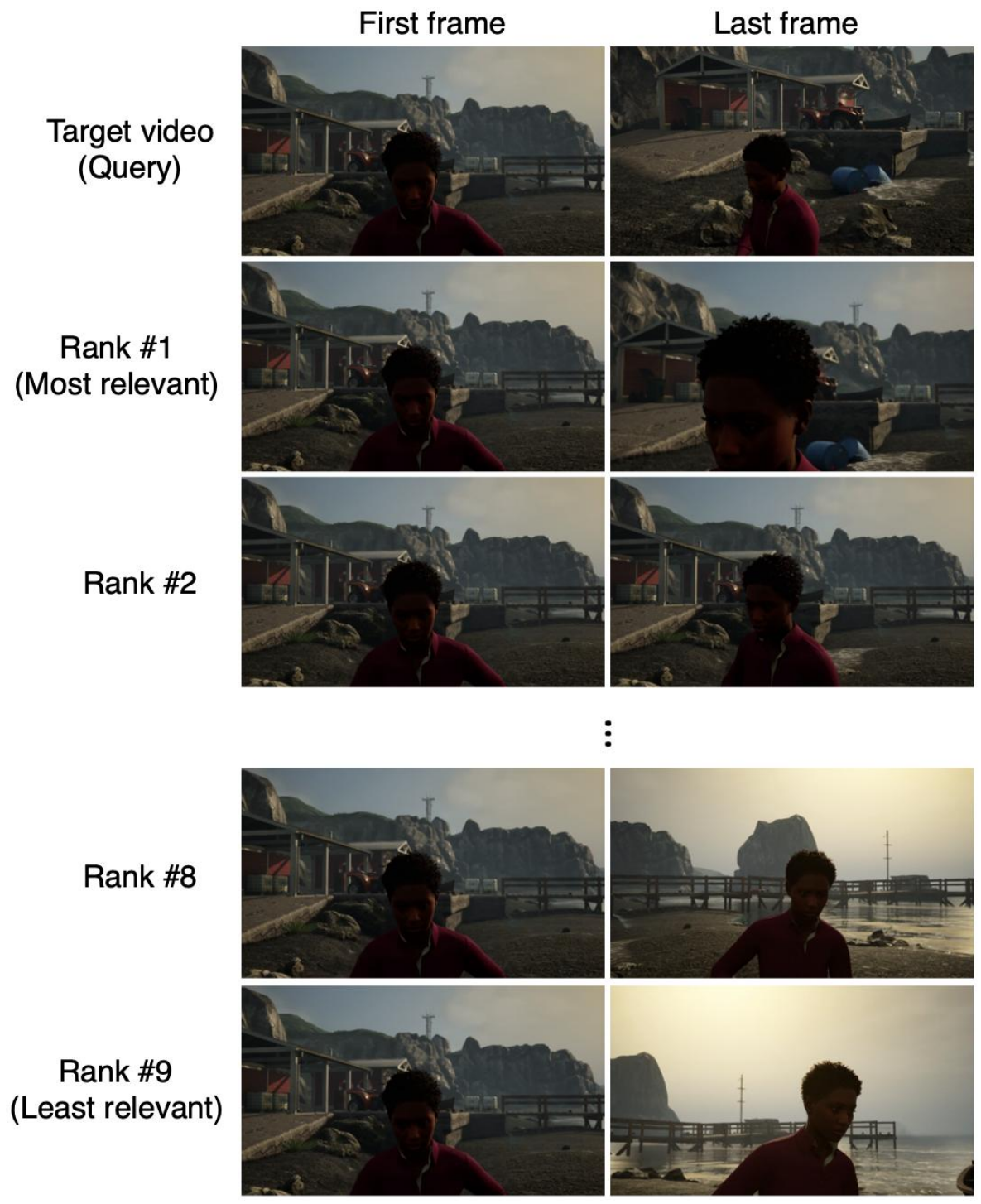}
    \vspace{-5mm}
    \caption{\textbf{Example video retrieval results using Alg.~\ref{alg:video-fov-retrieval}.}}
    \label{fig: retrieval_novel_view_synthesis}
\end{minipage}
\hfill
\begin{minipage}[t]{0.49\columnwidth}
    \centering
    \includegraphics[width=\linewidth]{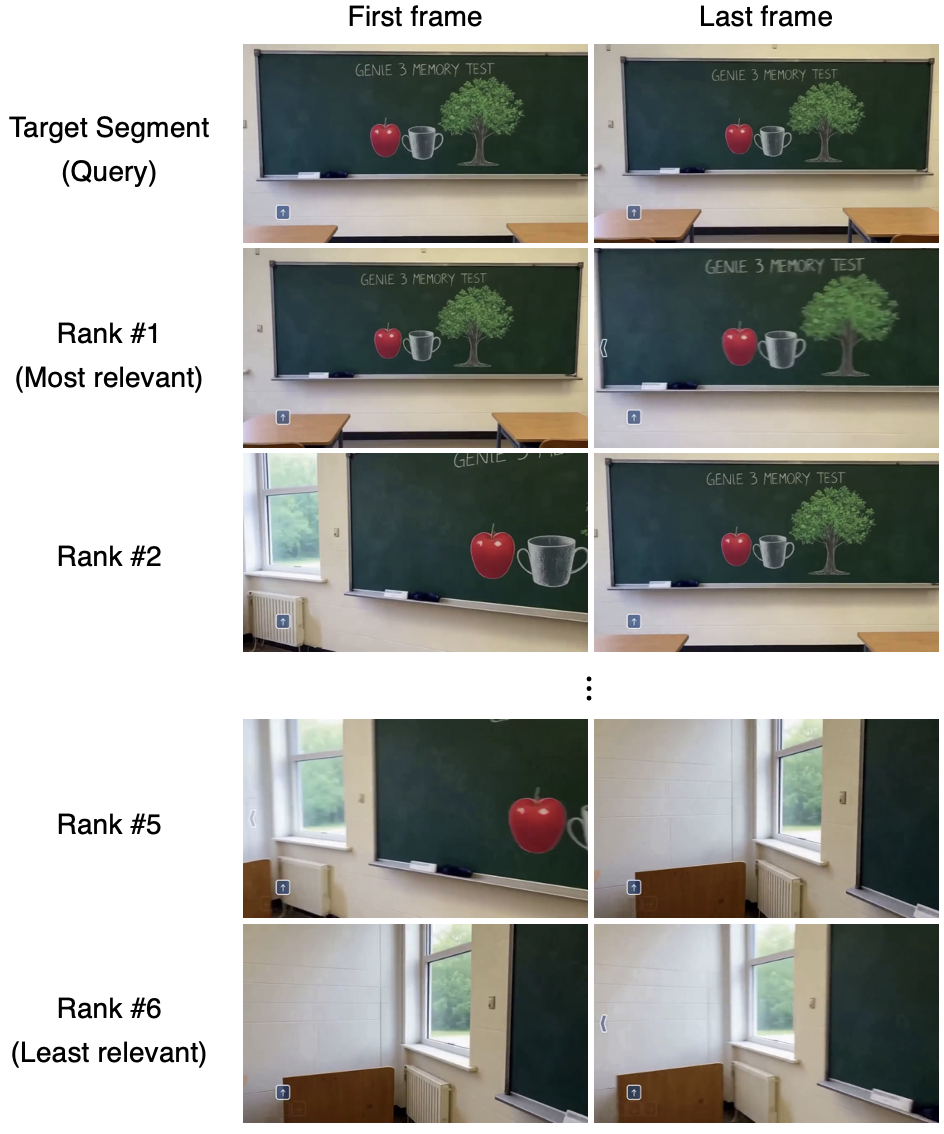}
    \caption{\textbf{Example video retrieval results using Alg.~\ref{alg:dino_segment_ranking}.}}
    \label{fig: retrieval_long_video_editing}
\end{minipage}

\end{figure}

Example retrieval results using the proposed VideoFOV algorithm are shown in Fig.~\ref{fig: retrieval_novel_view_synthesis}. The top row displays the target (query) video, while subsequent rows present retrieved videos ranked by descending relevance.

\subsection{Retrieval for Text-guided Long Video Editing}
For text-guided long video editing, retrieval is performed by selecting previously edited segments whose corresponding source segments are most visually similar to the current input segment.
Segment-level features are computed by extracting DINOv2~\cite{oquab2023dinov2} embeddings for all frames within a segment and averaging them into a single descriptor.
Cosine similarity between descriptors is then used to rank relevance.
To maintain temporal coherence between consecutive segments in the output, the most recently edited segment is always included in the retrieved set.
The full procedure is provided in Alg.~\ref{alg:dino_segment_ranking}. Example retrieval results are shown in Fig.~\ref{fig: retrieval_long_video_editing}, where semantically similar segments consistently rank highest.

\begin{table}[!t]
    \centering
    \definecolor{Gray}{gray}{0.9}
    \scriptsize
    \setlength{\tabcolsep}{3.5pt}
    \renewcommand{\arraystretch}{1.15}
    \begin{adjustbox}{max width=\columnwidth}
    \begin{tabular}{lccc}
        \toprule
         & \makecell{PSNR($\uparrow$)}
         & \makecell{SSIM($\uparrow$)}
         & \makecell{LPIPS($\downarrow$)} \\
        \midrule
        TrajectoryCrafter & 16.10 & 0.4353 & 0.3715 \\
        ReCamMaster & 16.40 & 0.4367 & 0.3895 \\
        \rowcolor{Gray}
        Memory-V2V (Top-1) & 17.15 & 0.4509 & 0.3483 \\ 
        \rowcolor{Gray}
        Memory-V2V (Top-3) & \textbf{18.05} & \textbf{0.4767} & \textbf{0.2973} \\
        \bottomrule
    \end{tabular}
    \end{adjustbox}
    
    \vspace{2mm}
    
    \caption{
    \textbf{Quantitative comparison results for NVS.}
    Memory-V2V achieves the best performance across all metrics, and Top-$3$ retrieval further improves reconstruction quality.
    }
    \label{tab:quantitative_result_nvs}
\end{table}

\begin{figure}[!t]
    \centering
    \includegraphics[width=0.9\columnwidth]
    {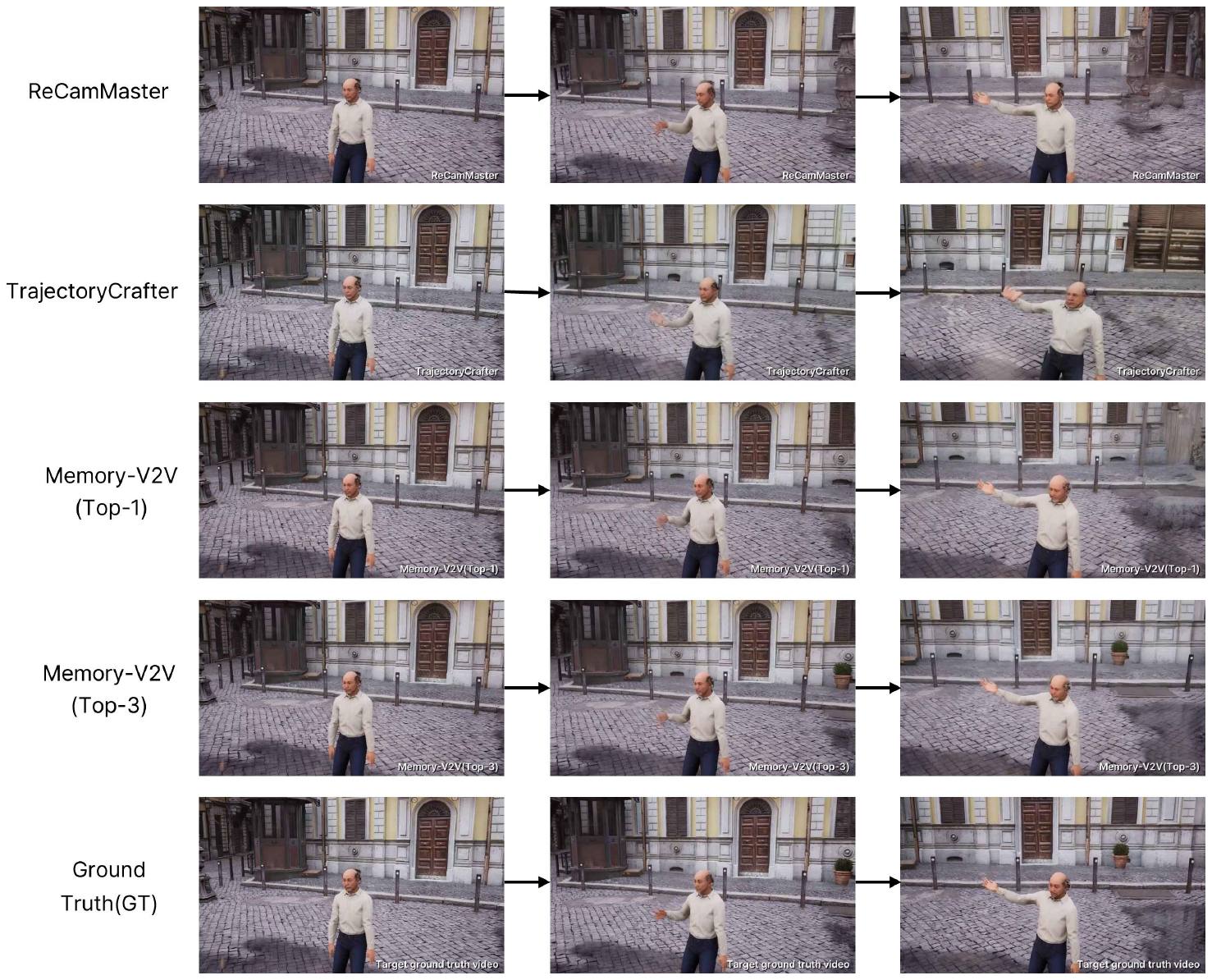}
    \caption{
    \textbf{Multi-view consistency with ground-truth multi-view videos.}
    Memory-V2V more faithfully reconstructs the target-view content and follows the target camera trajectory than prior methods. Top-$3$ retrieval further improves the reconstruction quality, particularly in regions that are weakly observed by the conditioning view.
    }
    \label{fig: nvs_gt}
\end{figure}

\section{Multi-view consistency with GT multi-view videos}
\label{sec: multi_view_consistency}
We conducted experiments on 50 validation scenes from the MultiCamVideo dataset, where each scene contains 10 synchronized views with ground-truth cameras. For baselines, we randomly selected a target camera and used another random ground-truth view as conditioning, then compared the generated result with the target-view ground-truth video. For our method, we used the same setup but retrieved memory videos using VideoFOV retrieval. We computed frame-wise PSNR, SSIM, and LPIPS against the target-view ground-truth in both Top-1 and Top-3 retrieval settings. Memory-V2V significantly outperforms baselines across all metrics (Tab.~\ref{tab:quantitative_result_nvs}), more faithfully reconstructs the left/middle regions, and better follows the target camera trajectory (Fig.~\ref{fig: nvs_gt}). Top-3 retrieval further improves both the metrics and the right region in Fig.~\ref{fig: nvs_gt}, confirming that additional memory enhances 3D fidelity, not only relative consistency.

\section{Additional Training and Inference Details}
\label{sec: training_inference_detail}
\subsection{Video Novel View Synthesis}

\subsubsection{Positional embedding (RoPE) design.}
Video diffusion models typically embed the spatiotemporal structure of videos using 3D Rotary Position Embeddings (3D-RoPE), where each token is assigned a positional encoding derived from its spatial coordinates $(x, y)$ and its temporal frame index $t$. Since all frames share the same spatial resolution, the spatial RoPE indices remain fixed across the video. In contrast, the temporal RoPE indices are bounded by the maximum video length seen during training. When the model is required to generate videos longer than this training horizon or incorporate additional conditioning videos during multi-turn editing, it must extrapolate to unseen temporal positions, which often leads to temporal drift or inconsistencies.

To resolve this issue, we first formalize the three types of videos involved in multi-turn video editing. The \emph{target video} is the video being generated in the current iteration and has temporal length $T$. The \emph{user-input video} is an externally provided video used to guide generation. The \emph{memory videos} consist of previously generated videos that are cached and reused as conditioning inputs. Unlike the target or user-input videos, the number of memory videos may grow unboundedly across editing iterations, requiring a positional encoding scheme that can support an expanding conditioning set without encountering unseen RoPE indices.

To provide a stable positional structure across these heterogeneous video sources, we assign each category a disjoint range of temporal RoPE indices. For a target video of length $T$, we define
\begin{align}
    \boldsymbol{\tau}_{\text{tgt}} &= \{0, 1, \dots, T{-}1\}, \\
    \boldsymbol{\tau}_{\text{usr}} &= \{T, T+1, \dots, 2T{-}1\}, \\
    \boldsymbol{\tau}_{\text{mem}} &= \{2T, 2T+1, \dots, 3T{-}1\}.
\end{align}
Each token at spatial location $(x, y)$ and temporal index $t$ receives a positional encoding given by
\begin{equation}
    \mathrm{RoPE}(x, y, t) 
    = [\mathrm{RoPE}_x(x), \mathrm{RoPE}_y(y), \mathrm{RoPE}_t(t)],
\end{equation}
where $t$ is sampled from $\boldsymbol{\tau}_{\text{tgt}}$, $\boldsymbol{\tau}_{\text{usr}}$, or $\boldsymbol{\tau}_{\text{mem}}$ depending on the video type.

Because this multi-video type configuration does not naturally arise during training, we adopt a mixed training strategy that enables the model to correctly interpret and utilize the hierarchical RoPE layout during inference. When both the user-input and memory RoPE ranges are active, we perturb the memory-video tokens with Gaussian noise to prevent the model from overfitting to artifacts that may accumulate across iterations. Specifically, for a memory token $\mathbf{z}_{\text{mem}}$, we use
\begin{equation}
    \tilde{\mathbf{z}}_{\text{mem}} 
    = \mathbf{z}_{\text{mem}} + \epsilon,
    \qquad 
    \epsilon \sim \mathcal{N}(0, \sigma^{2} I).
\end{equation}
This encourages the model to prioritize the cleaner and more reliable user-input video. Complementarily, we disable the user-input RoPE range with probability $p$ during training, forcing the model to rely solely on $\boldsymbol{\tau}_{\text{mem}}$. This RoPE dropout enables memory videos to serve as an independent conditioning signal.

Through this formulation and training procedure, the model learns a coherent and scalable RoPE structure that remains stable even as memory videos accumulate. The resulting system supports multi-turn video generation by enabling controlled information flow from the user-input video to the target video while mitigating the risk of error propagation through the memory video sets.

\subsubsection{Camera conditioning strategy.}
To incorporate viewpoint geometry into the generative process, we condition the feature representations on the camera extrinsics associated with each video in the conditioning set. For a video $v$ consisting of $f$ frames, we denote its camera representation by
\[
\mathrm{cam}_v \in \mathbb{R}^{\,f \times (3 \times 4)},
\]
where each $3 \times 4$ block corresponds to the rigid transformation $\left[R \;|\; t\right]$ describing the rotation and translation of the camera for an individual frame. These parameters are then projected into the feature space through a fully connected layer $\mathcal{E}_c(\cdot)$.

In ReCamMaster~\cite{bai2025recammaster}, target camera $cam_t$ information is injected uniformly across all features. Let $F_o$ denote the output feature of the spatial attention layer and $F_i$ denote the input feature to the subsequent 3D self-attention layer. The baseline formulation applies a residual update of the form
\begin{equation}
    F_i = F_o + \mathcal{E}_c(\mathrm{cam}_t),
\end{equation}
implicitly assuming that all conditioning features are derived from a single video captured under a single camera trajectory.

However, within our memory-driven framework, each memory video is generated under its own distinct camera extrinsics. This means that the conditioning set does not share a unified viewpoint, and the model must interpret the geometric context of each video individually. To achieve this, we assign to every feature token the camera embedding corresponding to the specific video from which it originates. If $F_o^{(v)}$ denotes a feature coming from video $v$, and $\mathrm{cam}_v$ is the associated camera trajectory, the camera-aware conditioning becomes
\begin{equation}
    F_i^{(v)} = F_o^{(v)} + \mathcal{E}_c\!\left(\mathrm{cam}_v\right).
\end{equation}

By embedding camera trajectories on a per-video basis in this manner, the model becomes capable of distinguishing heterogeneous viewpoints present across the target, user-input, and memory videos. This explicit geometric conditioning improves the model’s ability to handle complex camera motion and rotation, enabling more stable and coherent viewpoint reasoning during multi-turn video generation.

\begin{figure}[!t]
    \centering
    \includegraphics[width=0.7\columnwidth]
    {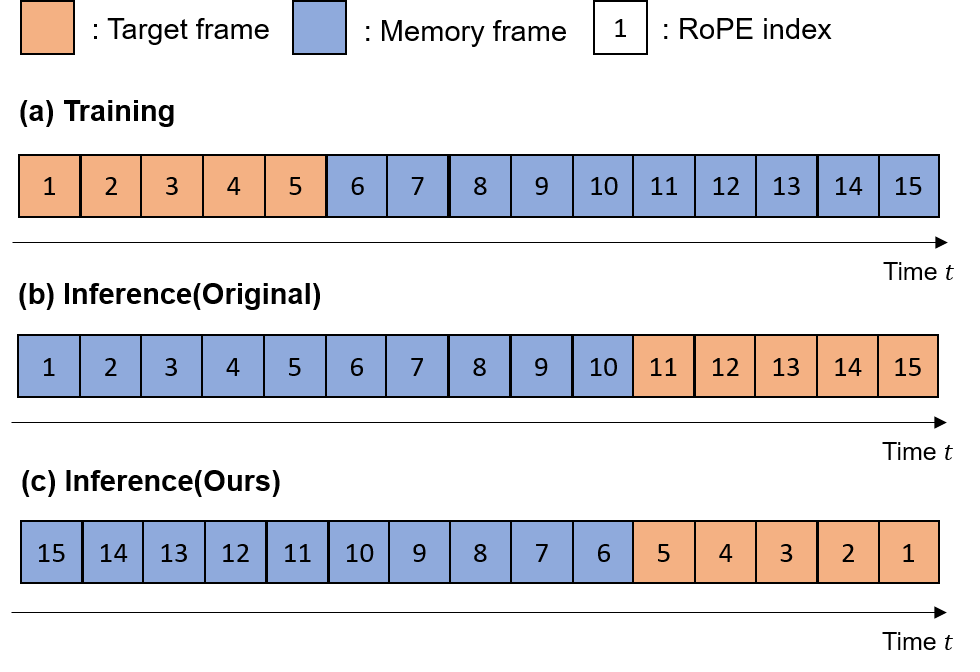}
    \caption{
    \textbf{RoPE assignments to mitigate train-inference gap.}
    }
    \label{fig: rope_inference}
    \vspace{-2mm}
\end{figure}

\subsection{Text-guided Long Video Editing}
\subsubsection{Dataset construction.}
Existing video editing datasets~\cite{zi2025se, bai2025ditto} are not suited for long video editing, as they contain only short clips with limited temporal context.
To enable training of Memory-V2V in this setting, we construct a long-form dataset by temporally extending video editing pairs from Señorita-2M~\cite{zi2025se}. Señorita-2M provides stable local editing pairs—where foreground edits minimally affect the background—but each clip is only 33 frames long, which is insufficient for long-horizon editing.
To address this limitation, we apply an off-the-shelf generative video extension model, FramePack~\cite{zhang2025packing}.
For each target clip, we retain the original 33 frames and generate 200 additional frames via forward temporal extension, resulting in a 233-frame sequence. 
The extended portion serves as memory during training; at each iteration, we randomly sample segments from this extended portion to function as memory-conditioning videos.

\subsubsection{Positional embedding (RoPE) design.}
Long-video editing follows the similar RoPE strategy used for video novel view synthesis. 
For each target segment of $T$ frames, we allocate disjoint temporal RoPE index ranges to distinguish token groups.
Specifically, the target segment is assigned as 
\[
\boldsymbol{\tau}_{\text{tgt}} = \{0, 1, \dots, T{-}1\},
\]
while the immediately preceding segment is assigned as
\[
\boldsymbol{\tau}_{\text{prev}} = \{T, T+1, \dots, 2T{-}1\},
\]
and other remaining segments are assigned as
\[
\boldsymbol{\tau}_{\text{mem}} = \{2T, 2T+1, \dots, 3T{-}1\}.
\]
This hierarchical indexing preserves strong continuity with the most recent segment while still providing broader historical context from earlier memory clips.

\subsubsection{Mitigating training-inference gap.}
During training, as shown in Fig.~\ref{fig: rope_inference} (a), the target video is temporally extended using a generative model; consequently, the target frames appear earlier in the sequence, while the extended (memory) frames appear later. RoPE indices are therefore assigned in chronological order from target to memory frames. However, during inference, the memory frames correspond to previously generated results, which naturally occur \emph{before} the target frames in time. If RoPE indices are assigned in the same forward order as shown in Fig.~\ref{fig: rope_inference} (b), this mismatch leads to a clear training-inference gap.
To eliminate this discrepancy, we reverse the RoPE assignment for memory frames during inference, as illustrated in Fig.~\ref{fig: rope_inference} (c). By flipping the RoPE index order, the positional structure of the conditioning sequence becomes consistent with the ordering used during training, thereby preventing the training-inference mismatch.

\begin{figure}[!t]
    \centering
    \vspace{-25mm}
    \includegraphics[width=1.0\columnwidth]
    {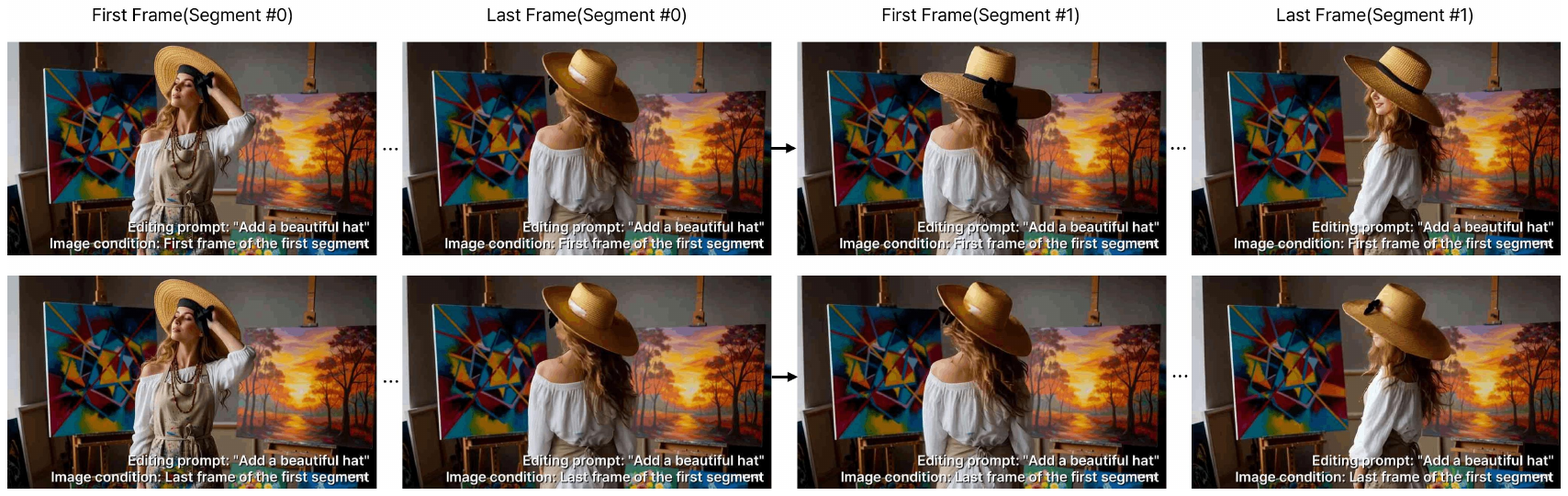}
    \vspace{-25mm}
    \caption{
    \textbf{Runway Aleph Results.}
    Runway Aleph struggles to maintain consistent edits under substantial viewpoint changes, particularly when the object turns around.
    }
    \label{fig: aleph}
\end{figure}

\begin{figure}[!t]
    \centering
    \vspace{-30mm}
    \includegraphics[width=1.0\columnwidth]
    {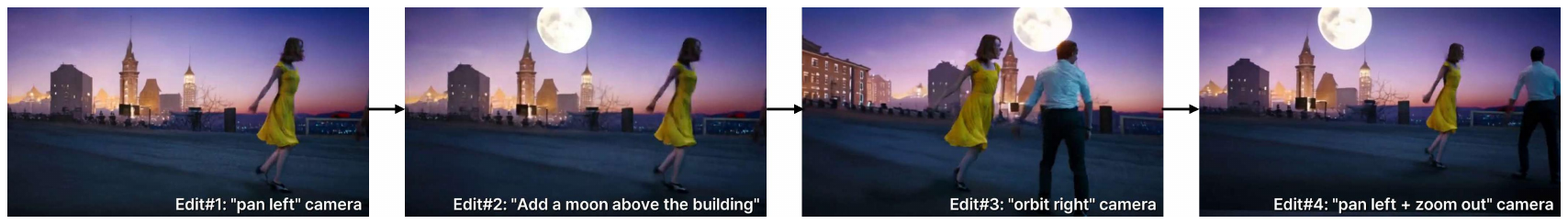}
    \vspace{-35mm}
    \caption{
    \textbf{Object-Level Consistency under Multi-View Editing.}
    A novel region is first revealed through camera motion, and a moon is inserted into the newly visible area. When subsequent camera trajectories revisit the same spatial region, Memory-V2V preserves the inserted moon at a consistent 3D location across viewpoints.
    }
    \label{fig: unified}
\end{figure}

\section{Comparison with a Recent Video-to-Video Model}
\label{sec: recent_v2v}
We compare Memory-V2V with Runway Aleph, a recent video-to-video editing model. Since Aleph edits videos of up to 5 seconds, we apply it to longer videos in a segment-wise manner, using frames from the previous segment as conditioning. With first-frame conditioning, Aleph better preserves the subject's frontal appearance and the edited attribute at the beginning of each segment. However, this often leads to inconsistencies when the viewpoint changes and back views appear. In contrast, last-frame conditioning provides better continuity near segment boundaries, but can propagate appearance mismatches from previous segments, such as inconsistent hats across the video (Fig.~\ref{fig: aleph}). Memory-V2V avoids this trade-off and produces more globally consistent editing results throughout long videos.

\section{Object-Level Consistency under Multi-View Editing}
\label{sec: object_level_consistency}
To further demonstrate the practical object-level consistency of Memory-V2V, Fig.~\ref{fig: unified} presents multi-view editing examples in which novel regions are first revealed through camera motion (e.g., pan left), and a new object (moon) is inserted into the newly visible region. We then apply \textit{orbit right} and \textit{pan left + zoom out} trajectories that revisit the same spatial area. The inserted moon consistently appears at the correct 3D location across multiple viewpoints. We also provide multi-segment text-editing results in the project page, consisting of 6 to 7 segments, in which objects disappear and reappear across editing turns while preserving their identity throughout the video. Although full unification of the two settings remains future work, they share the same memory conditioning architecture and differ only in the retrieval module: VideoFOV for multi-view editing and semantic similarity based retrieval for text-guided editing.

\section{Additional Qualitative Results}
\label{sec: additional_qual}
Additional results for multi-turn video novel view synthesis are presented in Fig.~\ref{fig: suppl_nvs_qual}, while further examples of text-guided long video editing are provided in Fig.~\ref{fig: suppl_edit_qual}. The arm--parrot intersection observed in Fig.~\ref{fig: suppl_edit_qual} reflects a limitation of the underlying base model rather than the proposed memory augmentation, as the artifact is also present in outputs without memory. We expect such interaction related failures to be substantially alleviated as the base model improves.
The remaining temporal jitter in the ``Change the hat into a more beautiful hat'' example shown on our project page mainly stems from imperfections in the synthetic training segments. Although we apply sharpness based filtering, minor artifacts may still accumulate over long sequences. We anticipate that more accurate video extension models will enable the construction of higher quality synthetic training data, which could further mitigate this issue.

\begin{figure}[!t]
    \centering
    \includegraphics[width=1.0\columnwidth]
    {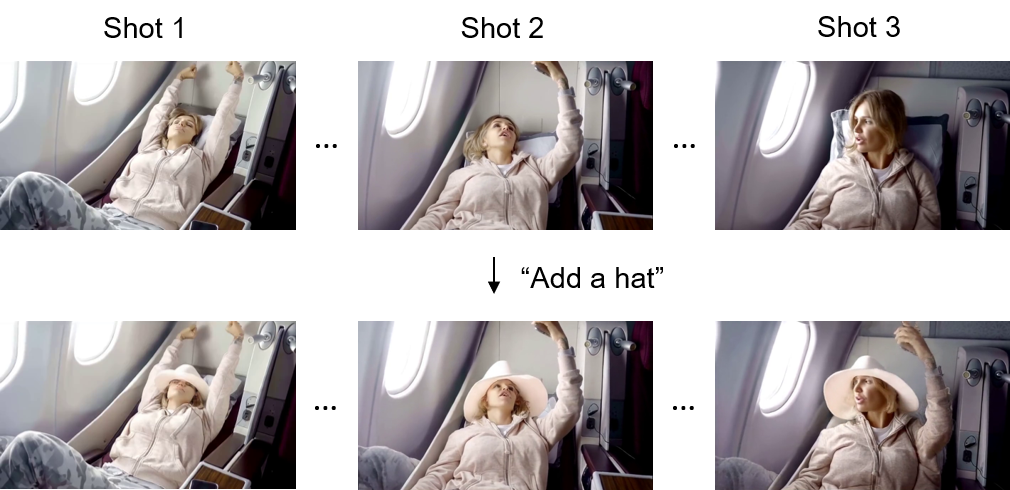}
    \caption{
    \textbf{Failure case.}
    Memory-V2V exhibits difficulties when the input long video contains multiple shots with large scene transitions.
    }
    \vspace{-5mm}
    \label{fig: multi_shot}
\end{figure}

\section{Additional Discussion}
\label{sec: discussion}

The dataset used to train Memory-V2V for text-guided long video editing consists exclusively of continuous, single-shot videos, without abrupt scene transitions. As a result, Memory-V2V struggles when applied to real long-form content containing multiple shots, where substantial visual discontinuities occur at shot boundaries (see Fig.~\ref{fig: multi_shot}). In such cases, the model may incorrectly propagate objects or textures from the preceding shot into the next (e.g., a hand or accessory reappearing), even if the high-level semantics remain similar.

Additionally, the target videos used during training are extended using a generative video model. These generated extensions exhibit mild flickering at the junction between real and synthesized segments due to temporal inconsistencies, slight changes in tone, or blur artifacts.
When such imperfect segments are stored in memory, these deviations accumulate over repeated denoising steps, especially in long sequences, ultimately leading to visible appearance drift.

We believe this limitation can be addressed by training with multi-shot datasets and higher-quality long-video data pairs, and we leave these directions to future work.
Moreover, to further enhance the interactivity of Memory-V2V, future work could integrate it with diffusion distillation or autoregressive generation frameworks \cite{yin2024one, yin2024improved, chen2024diffusion, yin2024slow}.

\begin{figure*}[!t]
    \centering
    \includegraphics[width=1.0\textwidth]
    {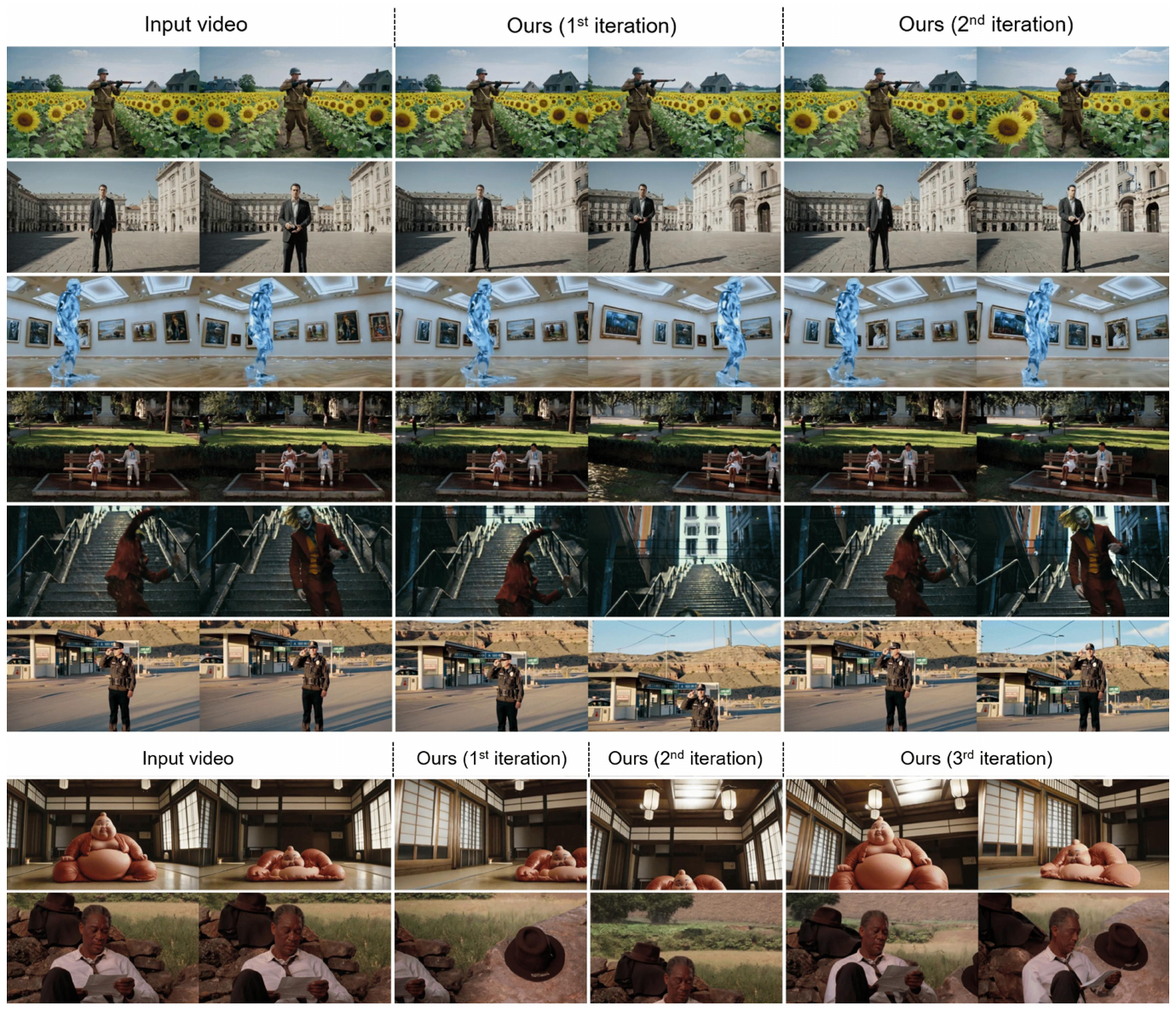}
    \caption{
    \textbf{Additional qualitative results for multi-turn video novel view synthesis.}
    Refer to our project page for video results.
    }
    \vspace{-5mm}
    \label{fig: suppl_nvs_qual}
\end{figure*}

\begin{figure*}[!t]
    \centering
    \includegraphics[width=1.0\textwidth]
    {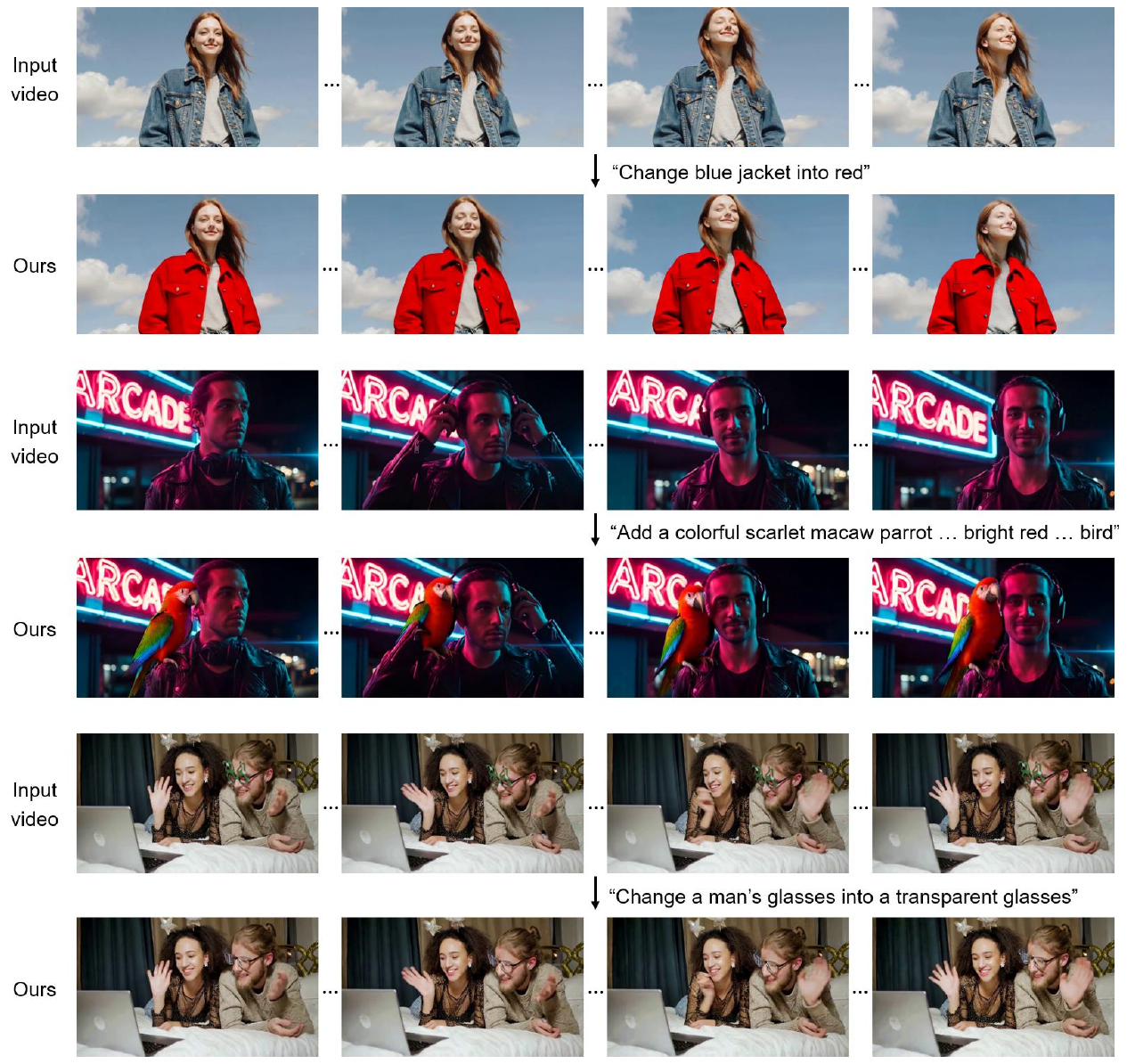}
    \caption{
    \textbf{Additional qualitative results for text-guided long video editing.}
     Refer to our project page for video results.
    }
    \label{fig: suppl_edit_qual}
\end{figure*}

\begin{algorithm}[t]
\caption{VideoFOV Retrieval: top-$k$ memory selection by spherical FOV overlap}
\label{alg:video-fov-retrieval}

\DontPrintSemicolon
\SetKwInOut{KwIn}{Input}
\SetKwInOut{KwOut}{Output}
\SetKwFunction{RelativePose}{RelativePose}
\SetKwFunction{InFOV}{InFOV}
\SetKwFunction{VideoFOV}{VideoFOV}

\KwIn{Target video $v^\ast$ with per-frame cameras $\{K_t^\ast, R_t^\ast, \mathbf{t}_t^\ast\}_{t=1}^{T^\ast}$; candidate memory videos $\{v_i\}_{i=1}^{N}$ with per-frame cameras $\{K_t^i, R_t^i, \mathbf{t}_t^i\}_{t=1}^{T_i}$; sphere radius $r$; number of sphere samples $M$; balance $\lambda\!\in\![0,1]$; top-$k$.}

\KwOut{$\tilde{\Omega}$ = top-$k$ videos ranked by FOV score.}

\BlankLine
\SetKwProg{Fn}{Function}{:}{end}
\Fn{\RelativePose{$R,\mathbf t, R_{\mathrm{ref}}, \mathbf t_{\mathrm{ref}}$}}{
  $R' \gets R_{\mathrm{ref}}^\top R$\;
  $\mathbf t' \gets R_{\mathrm{ref}}^\top(\mathbf t - \mathbf t_{\mathrm{ref}})$\;
  \KwRet $(R', \mathbf t')$\;
}

\Fn{\InFOV{$\mathbf p, K, R, \mathbf t$}}{
  $\mathbf x_c \gets R(\mathbf p - \mathbf t)$ \tcp*{World $\rightarrow$ camera}
  \If{$(\mathbf x_c)_z \le 0$}{\KwRet \textbf{false}}
  $\tilde{\mathbf u} \gets K\bigl(\mathbf x_c / (\mathbf x_c)_z\bigr)$ \tcp*{Project to pixels}
  \KwRet $\tilde{\mathbf u}$ inside image bounds\;
}

\Fn{\VideoFOV{$\{K_t,R_t,\mathbf t_t\}_{t=1}^{T},\; S=\{\mathbf p_m\}_{m=1}^M$}}{
  $\mathcal F \gets \varnothing$\;
  \For{$t \gets 1$ \KwTo $T$}{
    \For{$m \gets 1$ \KwTo $M$}{
      \If{\InFOV{$\mathbf p_m, K_t, R_t, \mathbf t_t$}}{
        $\mathcal F \gets \mathcal F \cup \{\mathbf p_m\}$\;
      }
    }
  }
  \KwRet $\mathcal F$\;
}

\BlankLine
\textbf{Reference the target’s first frame as origin}\;

$(R_{\mathrm{ref}}, \mathbf t_{\mathrm{ref}}) \gets (R_1^\ast, \mathbf t_1^\ast)$\;
Relativize all frames of $v^\ast$ and each $v_i$ via \RelativePose{}\;

Sample $S=\{\mathbf p_m\}_{m=1}^M$ uniformly on the sphere of radius $r$\;
$\mathcal F^\ast \gets$ \VideoFOV{$\{K_t^{\ast},R_t^{\ast},\mathbf t_t^{\ast}\}, S$}\;

\For{$i \gets 1$ \KwTo $N$}{
  $\mathcal F_i \gets$ \VideoFOV{$\{K_t^{i},R_t^{i},\mathbf t_t^{i}\}, S$}\;
  
  $s_{\mathrm{overlap}} \gets \dfrac{|\mathcal F^\ast \cap \mathcal F_i|}{|\mathcal F^\ast \cup \mathcal F_i|}$\;
  $s_{\mathrm{contain}} \gets \dfrac{|\mathcal F^\ast \cap \mathcal F_i|}{|\mathcal F^\ast|}$\;
  
  $s_i \gets \lambda\, s_{\mathrm{overlap}} + (1-\lambda)\, s_{\mathrm{contain}}$\;
}

\KwRet $\tilde{\Omega} \gets \text{top-}k$ videos by $s_i$ 

\end{algorithm}

\begin{algorithm}[t]
\caption{DINOv2-based Segment Similarity Ranking}
\label{alg:dino_segment_ranking}
\DontPrintSemicolon

\KwIn{
Target segment $X_{\mathrm{tar}}$ with frames $\{F_t^{\mathrm{tar}}\}_{t=1}^{T_{\mathrm{tar}}}$;\\
previous segments $\{X_k\}_{k=1}^N$ with frames $\{F_t^{(k)}\}_{t=1}^{T_k}$;\\
DINO backbone $f_\theta$; flag \texttt{enforce\_recent\_first}.
}
\KwOut{
Sorted indices $\pi$; similarity scores $\{\mathrm{sim}_k\}_{k=1}^N$.
}
\BlankLine

\textbf{Function} $\textsc{Descriptor}(X)$:\\
\Indp
Let $X$ contain frames $\{F_t\}_{t=1}^{T}$\;
$d \leftarrow \frac{1}{T} \sum_{t=1}^{T} f_\theta(F_t)$\;
\Return $d$\;
\Indm
\BlankLine

\textbf{Function} $\textsc{Sim}(d_i, d_j)$:\\
\Indp
$\mathrm{sim} \leftarrow 
\dfrac{\langle d_i, d_j \rangle}{\|d_i\|_2 \, \|d_j\|_2}$\;
\Return $\mathrm{sim}$\;
\Indm
\BlankLine

$d_{\mathrm{tar}} \leftarrow \textsc{Descriptor}(X_{\mathrm{tar}})$\;

\For{$k \leftarrow 1$ \KwTo $N$}{
    $d_k \leftarrow \textsc{Descriptor}(X_k)$\;
    $\mathrm{sim}_k \leftarrow \textsc{Sim}(d_{\mathrm{tar}}, d_k)$\;
}

$\pi \leftarrow \mathrm{argsort}(\{\mathrm{sim}_k\}, \text{descending})$\;

\If{\texttt{enforce\_recent\_first}}{
    Move index $N$ to the front of $\pi$\;
}

\Return $\pi,\ \{\mathrm{sim}_k\}_{k=1}^N$\;

\end{algorithm}

\clearpage

\end{appendix}
%
%
\bibliographystyle{splncs04}
\bibliography{main}
\end{document}